\documentclass{article}

\usepackage{PRIMEarxiv}

\errorcontextlines\maxdimen

\usepackage[utf8]{inputenc} % allow utf-8 input
\usepackage[T1]{fontenc}    % use 8-bit T1 fonts
\usepackage{url}            % simple URL typesetting
\usepackage{booktabs}       % professional-quality tables
\usepackage{nicefrac}       % compact symbols for 1/2, etc.
\usepackage{microtype}      % microtypography
\usepackage{fancyhdr}       % header
\usepackage{graphicx}
\usepackage{xcolor}
\usepackage{threeparttable}
\usepackage{subcaption}

\usepackage{lineno,hyperref}
\hypersetup{colorlinks, citecolor=blue, linkcolor=red, urlcolor=green}
\usepackage{amsmath,amsfonts,amssymb}
\usepackage{graphics}
\usepackage{bm}
\usepackage{multirow}

\usepackage{algorithm}
\usepackage{algpseudocode}

\usepackage{mathtools}
\usepackage{siunitx}
\DeclarePairedDelimiterXPP\BigOSI[2]%
  {\mathcal{O}}{(}{)}{}%
  {\SI{#1}{#2}}

\title{A Bayesian framework for discovering interpretable Lagrangian of dynamical systems from data}

\author{Tapas Tripura\\
  Department of Applied Mechanics\\
  Indian Institute of Technology Delhi\\
  Hauz Khas, 110016, India\\
  \texttt{tapas.t@am.iitd.ac.in} \\
  \And
  Souvik Chakraborty \\
  Department of Applied Mechanics\\
  Yardi School of Artificial Intelligence (ScAI)\\
  Indian Institute of Technology Delhi\\
  Hauz Khas, 110016, India\\
  \texttt{souvik@am.iitd.ac.in} \\
}

\begin{document}
\maketitle

\begin{abstract}
Learning and predicting the dynamics of physical systems requires a profound understanding of the underlying physical laws. Recent works on learning physical laws involve generalizing the equation discovery frameworks to the discovery of Hamiltonian and Lagrangian of physical systems. While the existing methods parameterize the Lagrangian using neural networks, we propose an alternate framework for learning interpretable Lagrangian descriptions of physical systems from limited data using the sparse Bayesian approach.
Unlike existing neural network-based approaches, the proposed approach (a) yields an interpretable description of Lagrangian, (b) exploits Bayesian learning to quantify the epistemic uncertainty due to limited data, (c) automates the distillation of Hamiltonian from the learned Lagrangian using Legendre transformation, and (d) provides ordinary (ODE) and partial differential equation (PDE) based descriptions of the observed systems. Six different examples involving both discrete and continuous system illustrates the efficacy of the proposed approach.
\end{abstract}

% keywords can be removed

\section{Introduction}
Modeling complex physical laws of dynamical systems using data-driven methods is an important area of research across various corners of science and engineering disciplines. Adoption of data-driven methods for system identification can be majorly found in the model updating of dynamical system \cite{adhikari2010distributed,rogers2020application,sengupta2023two}, design of optimal controllers \cite{roy2014robust,alibrandi2015optimal,patro2023kaimal}, digital twin \cite{worden2020digital,rahman2022leveraging,tripura2023probabilistic}, condition monitoring \cite{friswell2010structural,inturi2023integrated}, damage identification \cite{worden2009evidence,machado2017spectral,bhowmik2020robust}, remaining useful life prediction \cite{si2011remaining,mosallam2016data,zhu2020new}, data-driven reliability analysis \cite{khorshidi2015data,navaneeth2022surrogate,ye2023dynamic,mathpati2023mantra}, and surrogate modeling \cite{badawy2017hybrid,chakraborty2019surrogate,minh2020surrogate}.
Continual growth in pattern recognition and representation learning techniques has further given rise to intelligent data-driven algorithms that learn the interpretable mathematical representations of physical systems through the discovery of ordinary (ODE) \cite{bongard2007automated,schmidt2009distilling,brunton2016discovering,nayek2021spike,fuentes2021equation,tripura2023sparse} and partial differential equations (PDEs) \cite{rudy2017data,schaeffer2017sparse,chen2021physics,more2023bayesian,mathpati2023discovering}. While classical approaches to deriving underlying mathematical models were deeply rooted in the first principles and the deep understanding of the underlying physical systems, the modern data-driven methods simplified the model developments of complex systems where the physical knowledge is ambiguous, leaving a transformative impact on the growth of scientific and engineering discoveries. 
However, whether the learned interpretable ODE/PDE models conform to the underlying physical laws remains enigmatic. If the underlying physical laws are not satisfied by the discovered model, consequently, the generalization of the predictions over a long period of time for unseen environmental conditions will not be guaranteed. 
Therefore, it becomes more important to develop frameworks for representation learning of underlying physical symmetries rather than directly discovering the ODE/PDE descriptions \cite{baldi2017learning,wetzel2020discovering,liu2022machine,desai2022symmetry}. In this work, we propose an alternative method for discovering the Lagrangian of dynamical systems entirely from a single time-series observation of system states. Exploiting the recent advancements in sparse Bayesian learning (SBL) through sparsity-promoting spike-and-slab prior, we discover the exact interpretable Lagrangian of dynamical systems. 

Recently, attempts have been made using neural networks to learn the physical symmetries through Hamiltonian and Lagrangian of physical systems. Data-driven learning of Hamiltonian is evident in Hamiltonian Neural Networks (HNN) \cite{greydanus2019hamiltonian,toth2019hamiltonian} and Hamiltonian Graph Neural Networks (HGNN) \cite{sanchez2019hamiltonian}, where the neural network output is biased to satisfy the energy conservation, thereby satisfying the energy conservation law. 
Similarly, data-driven learning of Lagrangian can be traced to the Deep Lagrangian Network (DeLaN) \cite{lutter2019deep}, where the network is constrained to satisfy the Euler-Lagrangian equation, which was later improved to Lagrangian Neural Network (LNN) \cite{cranmer2020lagrangian}. Learning the Lagrangian of physical systems from the graph-based discretization of the physical domain is evident in Graph Lagrangian Neural Network (LGNN) \cite{bhattoo2022learning}. The trained networks exhibited energy conservation in these Lagrangian representation learning methods, therefore respecting governing physical laws.
For simplifying the learning process of Hamiltonian and Lagrangian through induction of explicit constraints, constrained Hamiltonian and Lagrangian networks were later proposed in \cite{finzi2020simplifying,gruver2022deconstructing}. 
Further extensions of Hamiltonian and Lagrangian neural networks for applications in robotic control, differentiable contact models, and video predictions can be found in \cite{duong2021hamiltonian,zhong2021extending,zhong2020unsupervised}.
Although these algorithms provided a way to alleviate the bounded processing capabilities of humans, they failed to extract reusable and transferable physics models of the underlying physical systems. These frameworks also require multiple-time histories of system trajectories to correctly identify the Lagrangian, which may sometimes be prohibitive from a practical point of view. In the absence of sufficient data, these neural networks violate energy conservation, thereby incorrectly learning the physical systems \cite{gruver2022deconstructing}. 

A step in the right direction is learning the energy-conserved parsimonious Lagrangian descriptions from data. In a more general framework, it is also desirable to learn the interpretable representations of Hamiltonian, the equations of motion, and the Lagrangian from a single time-series measurement of system states. 
We propose a Bayesian framework for identifying interpretable Lagrangian of physical systems from limited data. The salient features of the proposed framework can be encapsulated into the following points: (a) It learns the interpretable Lagrangian descriptions of physical systems using only a single observation of data, which is limited to a duration of one to two seconds. (b) Being Bayesian, it can quantify the epistemic uncertainty arising due to limited data. (c) It automates the distillation of interpretable Hamiltonians using the Legendre transformation on the learned Lagrangian densities. (d) It also provides the ODE/PDE descriptions of the underlying physical systems using the Euler-Lagrangian equation on identified Lagrangian.

In particular, this work will make the following contributions to the existing literature. 
(i) We propose a sparse Bayesian framework for automatic and accelerated learning of the exact interpretable form of Lagrangian from single observation data.
(ii) In a single framework, the framework can provide the analytical descriptions of Hamiltonian.
(iii) We show that the proposed framework can also provide the governing ODE/PDE descriptions of the underlying physical systems. 
(iv) We further show that the learned Lagrangian densities generalize to high-dimensional systems, which additionally possess the perpetual predictive ability. We rediscover the Lagrangian, Hamiltonian, and governing equations of motion of a series of examples, including systems represented by both linear and nonlinear ODE/PDEs, to demonstrate the success of our algorithm. 
The remainder of the paper is arranged as follows: in section \ref{sec:background}, the problem statement and a background on the sparse Bayesian regression using the spike-and-slab prior is given. In section \ref{sec:methods}, the proposed data-driven framework for discovering the Lagrangian from data is briefly presented. In section \ref{sec:numerical}, numerical experiments are undertaken to showcase the novelty of the proposed data-driven framework. In section \ref{sec:conclusion}, the contributions of the proposed framework are revisited, and finally, the paper is concluded.

\section{Background}\label{sec:background}

\subsection{Lagrangian and Euler-Lagrangian equation}\label{sec:problem}
We consider an $m$ dimensional physical system whose position at any time is tracked using position vector $\bm{r}_i$ for $i=1, \ldots, m$. The position vector is related to the generalized displacement and velocity coordinates $\bm{X}=\{X_i(t), \ldots, X_m(t) \}$ and $\dot{\bm{X}}=\{\dot{X}_1(t), \ldots, \dot{X}_m(t) \}$ through the transformations of the form $\bm{r}_j = \bm{r}_j(X_i(t), \ldots, X_m(t) )$ and $\dot{\bm{r}}_j = \dot{\bm{r}}_j(X_i(t), \ldots, X_m(t), \dot{X}_1(t), \ldots, \dot{X}_m(t) )$. 
Since the kinetic energy of the system $\mathcal{T}$ is a function of $\dot{\bm{r}}_i$, i.e., $\mathcal{T} = 0.5 \sum^m_{i=1} m_i \dot{\bm{r}}_i \cdot \dot{\bm{r}}_i$, the kinetic and potential energies of the physical system are denoted as $\mathcal{T}:= \mathcal{T}(X_1, \ldots, X_m, \dot{X}_1, \ldots, \dot{X}_m)$, and $\mathcal{V}:= \mathcal{V}(X_1, \ldots, X_m)$, respectively. This means that the potential energy $\mathcal{V}$ depends on the generalized displacement $X_i$, whereas the kinetic energy $\mathcal{T}$ can depend on both generalized displacement $X_i$ and velocity $\dot{X}_i$. 
From the Hamiltonian principle \cite{brizard2014introduction}, we have,
\begin{equation}\label{eq:differential}
    \int^{t_2}_{t_1} \left( \delta \mathcal{T} +\delta W \right) dt = 0,
\end{equation}
where $\delta W = \sum^m_{i=1} f^{ext}_{i} \cdot \delta \bm{r}_i$ is the virtual work done by all forces on the system, and $f^{ext}_{i}$ is the resultant force acting on the $i^{th}$ degree-of-freedom (DOF). The total virtual work done $\delta W$ can be decomposed into conservative and nonconservative components as $\delta W = \delta W^{c} + \delta W^{nc}$, where $\delta W^{c}=-\delta \mathcal{V}$, and $\delta W^{nc} = \sum^m_{i=1} Q_{i} \delta {X}_i$. Here, $Q_{i} = \sum^m_{i=1} f^{ext}_{i} \cdot (\partial \bm{r}_i / \partial X_i)$ is the generalized force.   
Using these results, we can rephrase the integral in Eq. \eqref{eq:differential} as,
\begin{equation}
    \int^{t_2}_{t_1} \biggl( \delta \mathcal{T} - \delta \mathcal{V} + \sum_{i=1}^m Q_i \delta X_i \biggr) dt = 0.
\end{equation}
At this stage, we introduce the Lagrangian $\mathcal{L}(\bm{X}, \dot{\bm{X}} ) \triangleq \mathcal{T}(\bm{X}, \dot{\bm{X}}) - \mathcal{V}(\bm{X})$ and write the above equation as,
\begin{equation}
    \int^{t_2}_{t_1} \left( \delta \mathcal{L}(\bm{X}, \dot{\bm{X}} ) + \sum_{i=1}^m Q_i \delta X_i \right) dt = \int^{t_2}_{t_1} \left( \sum^{m}_{i=1} \left[ \frac{\partial \mathcal{L}}{\partial X_i} \delta X_i + \frac{\partial \mathcal{L}}{\partial \dot{X}_i} \delta \dot{X}_i  + Q_i \delta X_i \right] \right) dt = 0.
\end{equation}
With integration by parts on the term $\int^{t_2}_{t_1} \sum^{m}_{i=1} ({\partial \mathcal{L}}/{\partial \dot{X}_i}) \delta \dot{X}_i dt$, we can obtain that,
\begin{equation}
    \int^{t_2}_{t_1} \left( \sum^{m}_{i=1} \left[ \frac{\partial \mathcal{L}}{\partial X_i}  - \frac{d}{dt} \left( \frac{\partial \mathcal{L}}{\partial \dot{X}_i} \right)  + Q_i  \right] \delta X_i \right) dt = 0.
\end{equation}
For the Hamiltonian principle to be zero, the terms inside the square bracket must vanish, i.e.,
\begin{equation}
    \frac{d}{dt} \left( \frac{\partial \mathcal{L}}{\partial \dot{X}_i} \right) - \frac{\partial \mathcal{L}}{\partial X_i} = Q_i; \;\; i=1,2, \ldots, m .
\end{equation}
The above equation is the Euler-Lagrange equation, also known as the Lagrangian equation of motion. If all the forces are conservative, then $Q_i = 0$, for $i=1,2, \ldots, m$, we obtain the Lagrangian equation of motion for nonconservative systems as, 
\begin{equation}\label{eq:lagrange_hom}
    \frac{d}{dt} \left( \frac{\partial \mathcal{L}}{\partial \dot{X}_i} \right) - \frac{\partial \mathcal{L}}{\partial X_i} = 0; \;\; i=1,2, \ldots, m .
\end{equation}
Given the temporal snapshot of the system states $\mathbf{X}=\{\bm{X}_i^{\top}, \ldots, \bm{X}_m^{\top} \}$ and $\dot{\bm{X}}=\{\dot{\bm{X}}_1^{\top}, \ldots, \dot{\bm{X}}_m^{\top} \}$, our aim is (I) to discover the exact interpretable mathematical form of the Lagrangian density $\mathcal{L}$ solely from the single time series observation, (II) to distill the associated Hamiltonian of the system, and (III) to derive the governing equations of motion of the underlying system.

\subsection{Sparse Bayesian regression using spike-and-slab prior}\label{sec:sbl}
Sparse regression is concerned with selecting a skewed subset of predictor variables from a design matrix in linear regression. In Bayesian learning, this is achieved by employing sparsity-inducing shrinkage prior \cite{mitchell1988bayesian,tipping2001sparse}. A well-known shrinkage model is the spike-and-slab prior with a point mass at 0 and a diffused slab distribution elsewhere \cite{nayek2021spike,tripura2023sparse}. 
The relation between the dependent and predictor variables in an inverse linear regression is expressed using the following equation:
\begin{equation}\label{eq:regression1}
	{\bm{Y}} = {\mathbf{L}}{\bm{\beta}} + \bm{\epsilon},
\end{equation}
where ${\bm{Y}} \in \mathbb{R}^N$ denotes the $N$-dimensional dependent vector, ${\mathbf{L}} \in \mathbb{R}^{N \times K}$ denotes the design matrix, ${\bm{\beta}} \in \mathbb{R}^{K}$ is the regression coefficient vector, and $\epsilon \in \mathbb{R}^N$ is the model mismatch error. The mismatch error $\epsilon$ is modeled as $i.i.d$ Gaussian random variable with zero mean and variance $\sigma^2_{obs}$. 
The posterior distribution of the weight vector ${\bm{\beta}}$ is obtained by applying the Bayes theorem,
\begin{equation}\label{eq:bayes}
    p\left( {\bm \beta, \sigma^2_{obs} |{\bm{Y} }} \right) \propto {p\left( {{\bm{Y}}|{\bm{\beta}, \sigma^2_{obs}} } \right) p\left( \bm \beta \right)} ,
\end{equation}
where $p\left( {\bm \beta, \sigma^2_{obs} |{\bm{Y}}} \right)$ is the posterior distribution of $\bm{\beta}$, $p\left( {{\bm{Y}}|{\bm{\beta}, \sigma^2_{obs}} } \right)$ is the likelihood function, and $p\left( \bm \beta \right)$ is the prior distribution of $\bm{\beta}$.
The likelihood distribution over $\bm{Y}$ is taken as, ${\bm{Y}}|{\bm{\beta }},{\sigma^2_{obs}} \sim \mathcal{N}\left( {{\mathbf{L}}{\bm{\beta}} ,{\sigma^2_{obs}}{{\bf{I}}_N}} \right)$, where ${\bf{I}}_{N}$ denotes the $\mathbb{R}^{N \times N}$ identity matrix.
For promoting sparsity in the weight vector $\bm{\beta}$, we utilize the discontinuous spike-and-slab (DSS) distribution, which is composed of a Dirac-delta function (denoting spike) and an independent Student’s-t distribution (denoting tail distribution), given as, 
\begin{equation}\label{dss}
	\begin{aligned}
	    p\left( {{\bm{\beta }}|{\bm{Z}}} \right) &= {p_{slab}}({\bm{\beta} _r}) \prod\limits_{k,{Z_k} = 0} {{p_{spike}}({\beta _k})}, \\
        &= \mathcal{N} \left( {{\bf{0}},{\sigma ^2_{obs}}{\vartheta _{slab}}{{\bf{I}}_{r}}} \right) \prod\limits_{k,{Z_k} = 0} {\delta _0},
	\end{aligned}
\end{equation}
where ${\delta _0}$ is the Dirac delta function centered at zero, and ${p_{slab}}({{\bm \beta} _r}) = \mathcal{N} \left( {{\bf{0}},{\sigma ^2_{obs}}{\vartheta _{slab}}{{\bf{I}}_{r}}} \right)$ is the slab distribution. Here, the new variable ${\bm{Z}} =\{Z_1, \ldots, Z_K\}$ is a latent indicator variable vector which classifies the distributions of the weights $\beta_k$ into slab and spike by taking a value of 1 if the weight corresponds to the slab component else takes a value of 0, ${\vartheta _{slab}}$ is the slab variance, and ${\bm{\beta}}_r \in \mathbb{R}^r$ is reduced regression coefficient vector composed from the elements of the weight vector ${\bm{\beta}}$ for which $Z_k=1$. 
The noise variance $\sigma^2_{obs}$ and the slab variance $\vartheta_{slab}$ are assigned the distributions $p({\sigma ^2_{obs}} ) = {\operatorname{IG}}( {{a _\sigma },{b _\sigma }} )$ and $p({\vartheta _{slab}} ) = {\operatorname{IG}}( {{a _\vartheta },{b _\vartheta }} )$, respectively. The latent variables ${Z_k}$ are assigned the distribution $p({Z_k}|{q} ) = {\operatorname{Bern}}( {q} );k = 1 \ldots K$ with the common hyperparameter $q$, where the hyperparameter $q$ is assigned the Beta distribution $p( {q} ) = {\operatorname{Beta}}( {{a _q},{b _q}} )$. IG is the Inverse-gamma distribution, Bern is the Bernoulli distribution, and Beta is the beta distribution.
The variables ${a _\vartheta }$, ${b _\vartheta }$, ${a _q }$, ${b _q }$, ${a _\sigma }$, and ${b _\sigma }$ are treated as deterministic hyperparameters. With the new variables $\bm{Z}$, $\vartheta_{slab}$, and $\sigma^2_{obs}$, the Bayesian inverse problem of estimating the posterior of $p\left( {\bm \beta, \sigma^2_{obs} |{\bm{Y} }} \right)$ becomes the hierarchical Bayesian inverse problem of estimating the following joint posterior distribution, 
\begin{equation}\label{eq:joint1}
    p\left( {{\bm{\beta }},{\bm{Z}},{\vartheta _{slab}},{\sigma ^2_{obs}},{q}|{\bm{Y}}} \right) \propto p\left( {{\bm{Y}}|{\bm{\beta }},{\sigma ^2_{obs}}} \right)p\left( {{\bm{\beta }}|{\bm{Z}},{\vartheta _{slab}},{\sigma ^2_{obs}}} \right)p\left( {{\bm{Z}}|{q}} \right)p\left( {{\sigma ^2_{obs}}} \right)p\left( {{\vartheta _{slab}}} \right)p\left( {q} \right),
\end{equation}
where $p( {{\bm{\beta }},{\bm{Z}},{\vartheta _{slab}},{\sigma ^2_{obs}},{q}|{\bm{Y}}} )$ is the joint posterior distribution of the random variables, $p( {{\bm{Y}}|{\bm{\beta }},{\sigma ^2_{obs}}} )$ is the likelihood distribution, $p( {{\bm{\beta }}|{\bm{Z}},{\vartheta _{slab}},{\sigma ^2_{obs}}} )$ is the prior for ${\bm{\beta}}$, $p( {{\bm{Z}}|{q}} )$ is the prior for ${\bm{Z}}$, $p( {{\vartheta _{slab}}} )$ is the prior for ${\vartheta}_{slab}$, $p( {{\sigma ^2_{obs}}} )$ is the prior for $\sigma^2_{obs}$, and $p( {{q}} )$ is the prior for $q$. 
Due to the discontinuous structure of the DSS prior, the direct sampling from the above joint distribution function is intractable. Therefore, the statistical inference from the posterior in Eq. \eqref{eq:joint1} is done using the Gibbs sampler \cite{geman1984stochastic}, which requires the conditional distributions of each random variable in the posterior distribution. The conditional distributions of the random variables for the above Bayesian inverse problem with spike-and-slab prior are found to be (see, e.g., \cite{nayek2021spike,tripura2023sparse} for a detailed discussion),
\begin{subequations}\label{eq:conditional}
    \begin{align}
    & p\left({{\bm{\beta }}_r}|{\bm{Y}},{\vartheta _{slab}},{\sigma ^{2}_{obs}} \right) = \mathcal N \left( {{{\bm{\mu}} _\beta },{{\mathbf{\Sigma}} _\beta }} \right), \label{gibbs_theta} \\
    & p\left({Z_k}|{\bm{Y}},{\vartheta _{slab}},{q} \right) = \operatorname{Bern}\left( \frac{{{q}}}{{{q} + \lambda \left( {1 - {q}} \right)}} \right), \;\; k=1,\ldots,K, \label{gibbs_z} \\
    & p\left({\sigma ^{2}_{obs}}|{\bm{Y}},{\bm{Z}},{\vartheta _{slab}} \right) = \operatorname{IG}\left( {{a _\sigma } + \frac{1}{2}N,{b _\sigma } + \frac{1}{2}\left( {{{\bm{Y}}^{\top}}{\bm{Y}} - {\bm{\mu}} _\beta^{\top} {\mathbf{\Sigma}} _\beta ^{- 1}{{\bm{\mu}} _\beta}} \right)} \right), \label{gibbs_sigma} \\
    & p\left({\vartheta _{slab}}|{\bm{\beta} },{\bm{Z}},{\sigma ^{2}_{obs}} \right) = \operatorname{IG}\left( {{a _\vartheta } + \frac{1}{2}{h_z},{b _\vartheta } + \frac{1}{2{\sigma ^2_{obs}}}{\bm{\beta }}_r^{\top}{\mathbf{I}}_{r}^{- 1}{{\bm{\beta }} _r}} \right), \label{gibbs_slab} \\
    & p\left({q}|{\bm{Z}} \right) = \operatorname{Beta}\left( {{a _q} + {h_z},{b _q} + K - {h_z}} \right), \label{gibbs_p0}
\end{align}
\end{subequations}
where the mean ${{\bm{\mu}} _\beta } = {\mathbf{\Sigma}}_\beta {\mathbf{L}}_r^{\top}{\bm{Y}}$, the covariance ${{\mathbf{\Sigma}} _\beta } = {\sigma ^{2}_{obs}}{\left( {{\mathbf{L}}_r^{\top}{{\mathbf{L}}_r} + \vartheta _{slab}^{- 1}{\mathbf{I}}_{r}^{- 1}} \right)^{ - 1}}$, $\lambda  = \frac{{p\left( {{\bm{Y}}|{Z_k} = 0,{{\bm{Z}}_{ - k}},{\vartheta _{slab}}} \right)}}{{p\left( {{\bm{Y}}|{Z_k} = 1,{{\bm{Z}}_{- k}},{\vartheta _{slab}}} \right)}} $ and ${{\mathbf{\Sigma}} _\beta }$, the parameters $\lambda$ and $h_z = \sum\nolimits_{k = 1}^K {{Z_k}}$. 
Here ${{\bm{Z}}_{-k}} \in \mathbb{R}^{K-1}$ denotes the latent variable vector consisting of all the elements except the $k^{th}$ component. 
The probabilities $p\left( {{\bm{Y}}|{\bm{Z}},{\vartheta _{slab}}} \right)$ indicating whether the latent variable $Z_k$ will take a value of 0 or 1, are given as,
\begin{equation}
    \begin{aligned}
        {p\left( {{\bm{Y}}|{Z_k} = 0,{{\bm{Z}}_{ - k}},{\vartheta _s}} \right)} & = \frac{{\Gamma \left( {{a _\sigma } + 0.5N} \right)b _\sigma ^{{a _\sigma }}}}{{\Gamma \left( {{a _\sigma }} \right){{\left( {2\pi } \right)}^{0.5N}}{{\left( {{b _\sigma } + 0.5{{\bm{Y}}^{\top}}{\bm{Y}}} \right)}^{\left( {{a _\sigma } + 0.5N} \right)}}}}, \\
        {p\left( {{\bm{Y}}|{Z_k} = 1,{{\bm{Z}}_{- k}},{\vartheta _{slab}}} \right)} & = \frac{{\Gamma \left( {{a _\sigma } + 0.5N} \right)b _\sigma ^{{a _\sigma }}{{\left( {\left| {{\mathbf{I}}_{r}^{- 1}} \right|\left| {{{\mathbf{\Sigma}} _\beta}} \right|} \right)}^{0.5}}}}{{\Gamma \left( {{a _\sigma }} \right){{\left( {2\pi } \right)}^{0.5N}}\vartheta _{slab}^{0.5{h_z}}{{\left( {{b _\sigma } + 0.5{{\bm{Y}}^{\top}}\left( {{{\mathbf{I}}_{N}} - {{\mathbf{L}}_r}{{\mathbf{\Sigma}} _\beta}{\mathbf{L}}_r^{\top}} \right){\bm{Y}}} \right)}^{\left( {{a _\sigma } + 0.5N} \right)}}}} .
    \end{aligned}
\end{equation}
Using the conditional distributions, the Gibbs sampler provided in Algorithm \ref{algo:gibbs} is performed to draw a sufficient number of samples so that the Markov Chain becomes stationary with respect to the posterior distribution in Eq. \eqref{eq:joint1}. 
\begin{algorithm}[ht!]
    \caption{Gibbs sampling of the posterior $p\left( {{\bm{\beta }},{\bm{Z}},{\vartheta _{slab}},{\sigma ^2_{obs}},{q}|{\bm{Y}}} \right)$}\label{algo:gibbs}
    \begin{algorithmic}[1]
	\Require{number of samples $n_s$ , burn-in size $n_b$, conditional densities (given in Eq. \eqref{eq:conditional}), and $j \gets 1$.}
        \State{Initial states of the Markov Chain $\bm{\theta}^{(*)}$, $\bm{Z}^{(*)}$, $\sigma^{2(*)}_{obs}$, $\vartheta_{slab}^{(*)}$, $q^{(*)}$.}
        \For{$i = 1, \ldots, n_s+n_b$}
            \State{// Sample model parameters: ${{\bm{\beta }}_r}^{(*)} \sim p\left(\cdot \mid {\bm{Y}},{\vartheta _{slab}^{(*)}},{\sigma ^{2(*)}_{obs}} \right)$.} \Comment{Eq. \eqref{gibbs_theta}}
            \State{// Sample latent variable vector: $Z_k^{(*)} \sim p\left(\bm{Z}_{-k}^{(*)} \mid {\bm{Y}},{\vartheta _{slab}^{(*)}},{q}^{(*)} \right)$, for $k=1,\ldots,K$.} \Comment{Eq. \eqref{gibbs_z}}
            \State{// Sample noise variance: ${\sigma ^{2(*)}_{obs}} \sim p\left(\cdot \mid {\bm{Y}},{\bm{Z}}^{(*)},{\vartheta _{slab}^{(*)}} \right)$.} \Comment{Eq. \eqref{gibbs_sigma}}
            \State{// Sample slab variance: ${\vartheta _{slab}^{(*)}} \sim p\left(\cdot \mid {\bm{\beta}}^{(*)},{\bm{Z}}^{(*)},{\sigma ^{2(*)}_{obs}} \right)$.} \Comment{Eq. \eqref{gibbs_slab}}
            \State{// Sample auxiliary variable: $q^{(*)} \sim p\left(\cdot \mid {\bm{Z}}^{(*)} \right)$.} \Comment{Eq. \eqref{gibbs_p0}}
            \If{$i > n_b$} 
                \State{// Save samples: $\bm{\theta}^{(j)} \gets \bm{\theta}^{(*)}$, $\bm{Z}^{(j)} \gets \bm{Z}^{(*)}$, $\sigma^{2(j)}_{obs} \gets \sigma^{2(*)}_{obs}$, $\vartheta_{slab}^{(j)} \gets \vartheta_{slab}^{(*)}$, $q^{(j)} \gets q^{(*)}$.}
                \State{$j \gets j+1$.}
            \EndIf
        \EndFor
        \Ensure{The Markov Chain: $\{\bm{\theta}^{(j)}, \bm{Z}^{(j)}, \sigma^{2(j)}_{obs}, \vartheta_{slab}^{(j)}, q^{(j)}\}^{(n_s)}_{j=1}$.}
    \end{algorithmic}
\end{algorithm}
The latent indicator variable vector $\bm{Z}^{(*)}$ is initialized by using a forward-backward grid search algorithm, given in Algorithm \ref{algo:search}. 
Since initial starting points may lie far away from the stationary distribution, few initial MCMC samples are discarded as burn-in samples. The inclusion of predictor variables in the final regression model is finalized using posterior inclusion probabilities (PIPs), which indicate the probability of occurrence of the corresponding predictor variable in predicting the dependent variable in an unseen scenario. The PIP$\triangleq p\left( {{Z_k} = 1|{\bm{Y}}} \right)$ is defined as \cite{nayek2021spike},
\begin{equation}\label{mpip}
    p\left( {{Z_k} = 1|{\bm{Y}}} \right) = \frac{1}{n_s}\sum\limits_{j = 1}^{n_s} {Z_k^j}; \;\; k = 1, \ldots ,K,
\end{equation}
where $n_s$ is the number of Gibbs iterations after discarding burn-in samples. The sample mean $\bm{\mu}_{\beta}$ and sample covariance $\mathbf{\Sigma}_{\beta}$ of the sparse regression coefficient vector $\bm{\beta}$ provides the uncertainty in the learned model arising due to the noisy and low data limit.

\section{Learning of Lagrangian of physical systems from state measurement}\label{sec:methods}
In this section, we introduce the sparse Bayesian Lagrangian identification algorithm for learning the exact interpretable forms of Lagrangian of physical systems from limited data. The algorithm automates the learning of the interpretable forms of Lagrangian from field observation data by leveraging the concepts of sparse Bayesian regression \cite{nayek2021spike,tripura2023sparse}. 
For learning the Lagrangian from data, the observations of displacement and velocity states of a physical system are collected for some duration $T$ at a discrete time step $\Delta t$. If $N$ is the length of temporal observations of system states, then we denote the measurements as $\mathbf{X} = [\bm{X}^{\top}_1, \ldots, \bm{X}^{\top}_m]$ and $\dot{\mathbf{X}} = [\dot{\bm{X}}^{\top}_1, \ldots, \dot{\bm{X}}^{\top}_m]$.

To be able to learn the interpretable forms of Lagrangian, the proposed algorithm expresses the Lagrangian $\mathcal{L}(\mathbf{X}, \dot{\mathbf{X}} )$ as a weighted linear superposition of certain linear and nonlinear basis functions. These basis functions effectively represent the various forms of kinetic and potential energies and are functions of state observation data. 
Assuming that there are a total of $K$-basis functions, we express the Lagrangian $\mathcal{L}$ for $i^{th}$ degree-of-freedom (DOF) as,
\begin{equation}\label{eq:lagrange_lib}
    \begin{aligned}
        \mathcal{L}_{i}(\mathbf{X}, \dot{\mathbf{X}} ) &= \beta_{i_1} f_1(\mathbf{X}, \dot{\mathbf{X}} ) + \beta_{i_2} f_2(\mathbf{X}, \dot{\mathbf{X}} ) + \ldots + \beta_{i_K} f_K(\mathbf{X}, \dot{\mathbf{X}} ) , \\
        & = \mathbf{L}(\mathbf{X}, \dot{\mathbf{X}}) \bm{\beta}_{i},
    \end{aligned}
\end{equation}
where $\mathbf{L}(\mathbf{X}, \dot{\mathbf{X}}) \in \mathbb{R}^{N \times K}$ is the design matrix containing all the possible choices of pre-selected basis functions $f_k(\mathbf{X}, \dot{\mathbf{X}})$, and $\bm{\beta}_{i}$ are the parameters of the Lagrangian of $i^{th}$-DOF. 
The functions $f_i(\mathbf{X}, \dot{\mathbf{X}})$ are mapping of each column of the observations matrices $\mathbf{X}$ and $\dot{\mathbf{X}}$. A demonstration of the design matrix employed in this work is given below, 
\begin{equation}\label{eq:library}
    \mathbf{L}(\mathbf{X}, \dot{\mathbf{X}}) = \left[ {\begin{array}{*{20}{c}}
            {\bf{1}} & {{{\rm P}^{\rho}} ({\mathbf{X}}, \dot{\mathbf{X}})} & {\mathop{\rm Hm.}({\mathbf{X}}, \dot{\mathbf{X}})} & {{{\rm P}^{\rho}}(\vert {\mathbf{X}}- {\mathbf{X}^{\prime}} \vert )} & {{{\rm P}^{\rho}}(\vert {\dot{\mathbf{X}}}- \dot{\mathbf{X}}^{\prime} \vert )} & {\mathop{\rm Hm.}(\vert {\mathbf{X}}- {\mathbf{X}^{\prime}} \vert )} & {\mathop{\rm Hm.}(\vert \dot{\mathbf{X}}- \dot{\mathbf{X}}^{\prime} \vert )}
    \end{array}} \right],
\end{equation}
where ${{{\rm P}^{\rho}} (\cdot,\cdot)}$ is the polynomial function, $\rho$ is the degree of polynomials, and ${\mathop{\rm Hm.}(\cdot, \cdot)}$ denote the family of harmonic functions like sine and cosine. In the remainder of the discussion, we drop the arguments of the $\mathcal{L}(\cdot, \cdot)$ and $\mathbf{L}(\cdot, \cdot)$. 

With a model mismatch error $\bm{\epsilon}$, the problem of identification of exact functions from $\mathbf{L}$ can be formulated as Eq. \eqref{eq:regression1}. However, since we do not have the knowledge of Lagrangian, we use the Euler-Lagrangian equation in Eq. \eqref{eq:lagrange_hom} to bias the discovered Lagrangian functions to obey the optimal path. This is done by substituting $\mathcal{L}_{i}(\mathbf{X}, \dot{\mathbf{X}} ) = \mathbf{L}(\mathbf{X}, \dot{\mathbf{X}}) \bm{\beta}_{i}$ in to Eq. \eqref{eq:lagrange_hom}, which yields,
\begin{equation}\label{eq:lagrange_motion}
    \left[ \frac{d}{d t} \left( \frac{\partial \mathbf{L}}{\partial \dot{X}_i} \right) - \frac{\partial \mathbf{L}}{\partial {X}_i} \right]  \bm{\beta}_i = \bm{0}, \; i=1,2, \ldots, m .
\end{equation}
The above equation is rephrased as, $\bm{0} = \hat{\mathbf{L}}\bm{\beta}_i + {\bm{\epsilon}}, \; i=1,2, \ldots, m$, where $\hat{\mathbf{L}} \in \mathbb{R}^{N \times K}$ denotes the terms within the square bracket in Eq. \eqref{eq:lagrange_motion}. We refer to this dictionary matrix as the Euler-Lagrange library, obtained by operating the Euler-Lagrangian operator ``$({d}/{dt})({\partial }/{\partial \dot{X}_i} ) - {\partial}/{\partial {X}_i}$" on the Lagrange library $\mathbf{L} \in \mathbb{R}^{N \times K}$. 
\begin{figure}[!t]
  \centering
  \includegraphics[width=\textwidth]{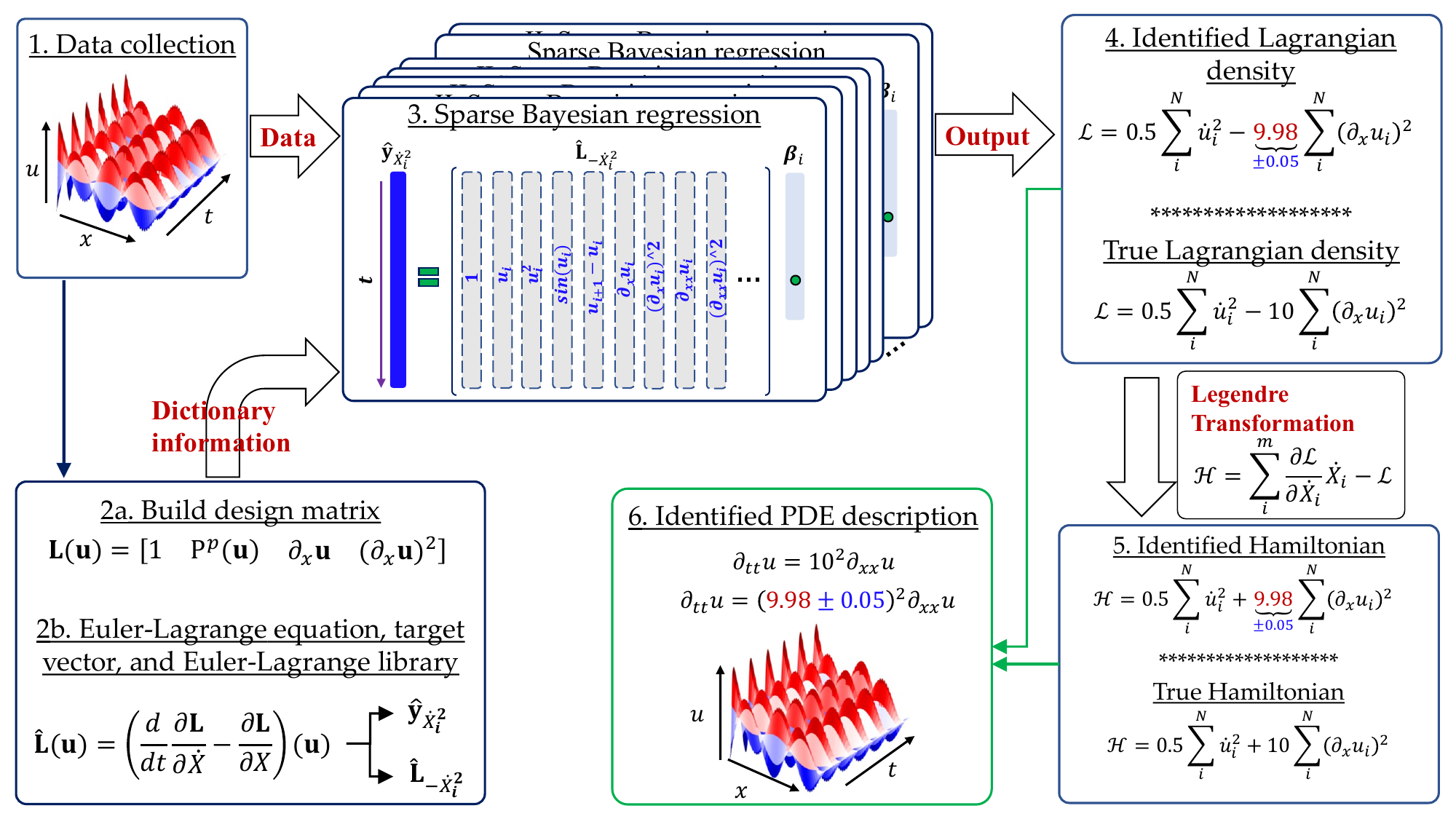}
  \caption{Proposed Bayesian framework for learning interpretable Lagrangian, illustrated through identification of the Euler-Barnoulli beam equation from limited data. Existing neural network-based approaches fail to provide a reusable Lagrangian description. On the contrary, the proposed framework learns an interpretable Lagrangian that is reusable and transferable. First, data are collected as snapshots at discrete time intervals and then transformed into a design matrix composed of certain candidate functions. The Euler-Lagrangian operator is applied to the design matrix to obtain the necessary components of the regression equation. Finally, sparse Bayesian regression is performed to identify the active candidate functions. The framework also provides the Hamiltonian, ODE/PDE descriptions, and epistemic uncertainty in the learned Lagrangian model.}
\end{figure}
However, since the target vector contains no information, it yields a trivial solution $\bm{\beta}_i=0$ for $i=1,\ldots,m$. To seek a non-trivial solution for the model parameters ${\bm{\beta}}_i$, we first locate the column index of the basis function $\dot{X}^2_i$ from $\mathbf{L}$ and then use the corresponding column vector from the Euler-Lagrange library $\hat{\mathbf{L}}$, denoted as $\hat{\bm{y}}_{\dot{X}_i^2}$, as the target vector, yielding the following regression,
\begin{equation}\label{eq:motion_opt2}
    \hat{\bm{y}}_{\dot{X}_i^2} = \hat{\mathbf{L}}_{-\dot{X}_i^2} {\bm{\beta}_i} + {\bm{\epsilon}} , \; i=1,2, \ldots, m ,
\end{equation}
where $\hat{\mathbf{L}}_{-\dot{X}_i^2} \in \mathbb{R}^{N \times (K-1)}$ is the reduced Euler-Lagrange library with the vector $\hat{\bm{y}}_{\dot{X}_i^2}$ removed from $\hat{\mathbf{L}}$.
The exact interpretable forms of Lagrangian can be identified from the field measurement data by solving the Eq. \eqref{eq:motion_opt2} using the sparse Bayesian regression discussed in section \ref{sec:sbl}. Note that it requires an $m$ distinct search to obtain the sparse matrix $\boldsymbol{\beta} = [\bm{\beta}_1, \ldots, \bm{\beta}_m]$, where $\bm{\beta}_i$ denote the sparse Lagrangian parameter vector associated with $i^{th}$-DOF.
Once the parameter vector $\bm{\beta}_i$ is obtained, the exact analytical form of the Lagrangian is obtained by augmenting the function $\dot{X}_i^2$ to the discovered Lagrangian, i.e.,
\begin{equation}\label{eq:lagrangian}
    \mathcal{L}_{i} (\mathbf{X}, \dot{\mathbf{X}} ) = {\mathbf{L}}_{-\dot{X}_i^2} \bm{\beta}_i + {\dot{X}_i^2}, \; i=1,2, \ldots, m .
\end{equation}
Finally, the Lagrangian of an $n$ DOF system is determined by performing an algebraic summation of the Lagrangian of all the DOFs, i.e., $\mathcal{L}(\mathbf{X}, \dot{\mathbf{X}} ) = \sum^{m}_{i} \mathcal{L}_{i} (\mathbf{X}, \dot{\mathbf{X}} )$. However, the Lagrangian will be normalized with respect to the coefficient of the basis function $\dot{\bm{X}}_i^2$. The implementation is shown in the Algorithm \ref{algo:lagrange}. Once the Lagrangian of a physical system is learned, the Hamiltonian of the system is derived from the Lagrangian using the Legendre transformation as,
\begin{equation}\label{eq:hamiltonian}
    \mathcal{H}(\mathbf{X}, \dot{\mathbf{X}}) = \sum^m_{i=1} {\partial _{\dot{X}_{i}}} \mathcal{L}(\mathbf{X}, \dot{\mathbf{X}}) \dot{X}_{i} - \mathcal{L}(\mathbf{X}, \dot{\mathbf{X}}) ,
\end{equation}
where ${\partial _{\dot{X}_{i}}}$ denotes the partial derivative with respect to the state ${\dot{X}_{i}}$ and $\mathcal{H}(\cdot, \cdot)$ is the Hamiltonian of the physical system. The governing equation of motions of a physical system can also be obtained from the same sparse coefficient vector ${\bm{\beta}}_i$ as, 
$\hat{\bm{y}}_{\dot{X}_i^2} - \hat{\mathbf{L}}_{-\dot{X}_i^2} {\bm{\beta}_i} =0, \; i=1,2, \ldots, m$.
As opposed to the equation discovery and Lagrangian neural network frameworks, where the physical laws are derived either in the form of equations of motion or Lagrangian, the proposed framework provides a single framework for discovering exact analytical descriptions of the Lagrangian and governing equations of motion.

\begin{algorithm}[ht!]
    \caption{Sparse Bayesian identification of Lagrangian}\label{algo:lagrange}
    \begin{algorithmic}[1]
    \Require{Observations $\mathbf{X} \in \mathbb{R}^{N \times m}$, $\dot{\mathbf{X}} \in \mathbb{R}^{N \times m}$, and hyperparameters.}
        \State{Construct the dictionary $\mathbf{L}(\mathbf{X}, \dot{\mathbf{X}})$.}\Comment{Eq. \eqref{eq:library}}
        \For{$i=1,\ldots,m$}
            \State{Compute the partial derivative of $\mathbf{L}$ with respect to $X_i$: $\partial_{X_i} \mathbf{L}$.}
            \State{Compute the partial derivative of $\mathbf{L}$ with respect to $\dot{X}_i$: $(d/dt) \partial_{\dot{X}_i} \mathbf{L}$.}
            \State{Obtain the Euler-Lagrangian library $\hat{\mathbf{L}} = (d/dt)(\partial_{\dot{X}_i} \mathbf{L}) - \partial_{X_i} \mathbf{L}$.}
            \State{Extract the target vector $\bm{y}_{\dot{X}^2_i}$ and the design matrix $\hat{\bm{L}}_{\dot{X}^2_i}$.}
            \State{Construct the regression problem: $\hat{\bm{y}}_{\dot{X}_i^2} = \hat{\mathbf{L}}_{-\dot{X}_i^2} {\bm{\beta}_i} + {\bm{\epsilon}}$.} \Comment{Eq. \eqref{eq:motion_opt2}}
            \State{Perform the sparse Bayesian regression.} \Comment{Algorithm \ref{algo:gibbs}}
            \State{Obtain the Lagrangian $\mathcal{L}_{i}(\mathbf{X}, \dot{\mathbf{X}} )$.} \Comment{Eq. \eqref{eq:lagrangian}}
        \EndFor
    \Ensure{Lagrangian: $\mathcal{L}(\mathbf{X}, \dot{\mathbf{X}}) = \sum_i^m \mathcal{L}_{i}(\mathbf{X}, \dot{\mathbf{X}} )$.}
    \end{algorithmic}
\end{algorithm}

\begin{algorithm}[h!]
    \caption{Algorithms for latent indicator variable vector inilizatation}\label{algo:search}
    \begin{algorithmic}[1]
    \Require{Design matrix $\mathbf{L}$ and target vector $\bm{Y}$.}
    \State{Initiate forward grid search: $\bm{\beta} \gets ((\mathbf{L}^{\top} \mathbf{L})^{-1}\mathbf{L}^{\top}) \bm{Y}$, $\bm{\epsilon}_{0}^{f} \gets (\bm{Y}-\mathbf{L}\bm{\beta})$.}
    \For{$i=1,\ldots,K$}
        \If{$i == 0$}
        \State{Activate the first candidate function from $\mathbf{L}$ and compute MSE: $\bm{\epsilon}_{0}^{f} = \bm{Y}-\mathbf{L}_{(1)}\bm{\beta}_{(1)}$.}
        \State{If $\bm{\epsilon}_{0}^{f} \leq \bm{\epsilon}_{0}$, $Z_{0} \gets 1$, otherwise, $Z_{0} \gets 0$.}
        \Else
        \State{Find $\bm{\beta}_r$ and $\mathbf{L}_{r}$; consisting only those basis functions from $\mathbf{L}$ for which $Z_{k}$=1.}
        \State{Stack with current basis function, $\bm{\beta}_i$ and $\mathbf{L}_{i}$:  $\hat{\bm{\beta}}$=$\{\bm{\beta}_r, \bm{\beta}_i\}$ and $\hat{\mathbf{L}}$=$\{\mathbf{L}_r,\mathbf{L}_{i}\}$.}
        \State{Compute $\bm{\epsilon}_{i}^{f} = \bm{Y}-\mathbf{L}\bm{\beta}$. If $\bm{\epsilon}_{i}^{f} \leq \bm{\epsilon}_{i-1}^{f}$, $Z_{j} \gets 1$, otherwise, $Z_{j} \gets 0$.}
        \EndIf
    \EndFor
    \State{Initiate backward grid search: obtain $\bm{\theta}_r$ and $\mathbf{L}_r$ for which $Z_k$=1, and compute $\bm{\epsilon}_{0}^{b} = \bm{Y}-\mathbf{L}_{r}\bm{\theta}_{r}$.}
    \For{$j$ in $K$}
        \If{$Z_{(K-j)}$ == 1}
        \State{Remove $j^{th}$ candidate function and obtain $\mathbf{L}_{r(-j)}$ and $\bm{\theta}_{r(-j)}$.}
        \State{Estimate MSE: $\bm{\epsilon}_{j}^{b} = \bm{Y}-\mathbf{L}_{r(-j)} \bm{\theta}_{r(-j)}$.}
        \State{If $\bm{\epsilon}_{j+1}^{b} \leq \bm{\epsilon}_{j}^{b}$, $Z_{j} \gets 0$, otherwise, $Z_{j} \gets 1$.}
        \EndIf
    \EndFor
    \Ensure{The latent indicator variable vector $\bm{Z} \in \mathbb{R}^{K}$.}
    \end{algorithmic}
\end{algorithm}

\section{Numerical examples}\label{sec:numerical}
In this section, we undertake six examples and try to recover their actual analytical Lagrangian forms from simulated data. We also discover the underlying system's Hamiltonian and the governing equations of motion as a side product of the proposed Lagrangian discovery algorithm. The ability to generalize to unseen initial conditions, perpetual predictive capability, and generalization to high-dimensional systems are also investigated towards the end of the section. For each example, only a single time series of the system states are generated. 
Following the previous works on spike and slab, the hyperparameters are taken as $a_p$=0.1, $b_p$=1, $a_v$=0.5, $b_v$=0.5, $a_\sigma$=$10^{-4}$, $b_\sigma$=$10^{-4}$, $p_0^{(0)}$=0.1, $\vartheta^{(0)}$=10, and $\sigma ^{2(0)}$=var$(\bm{\epsilon})$, where $\bm{\epsilon}$ is the residual error from ordinary least-squares \cite{nayek2021spike,o2009review}. To initialize the vector ${\bm{Z}}^{(0)}$ a forward-backward search algorithm is devised, which activates the $k^{th}$ element of ${\bm{Z}}$ if it reduces the ordinary lease-squared error between the training data and the obtained mode (see Algorithm \ref{algo:search}). A Markov chain of length 5000 with 1000 burn-in samples is used for sparse Bayesian regression. Basis functions with PIP$>$0.5, i.e., probability more than 0.5, are included in the final model.

\begin{figure}[t!]
    \centering
    \includegraphics[width=\textwidth]{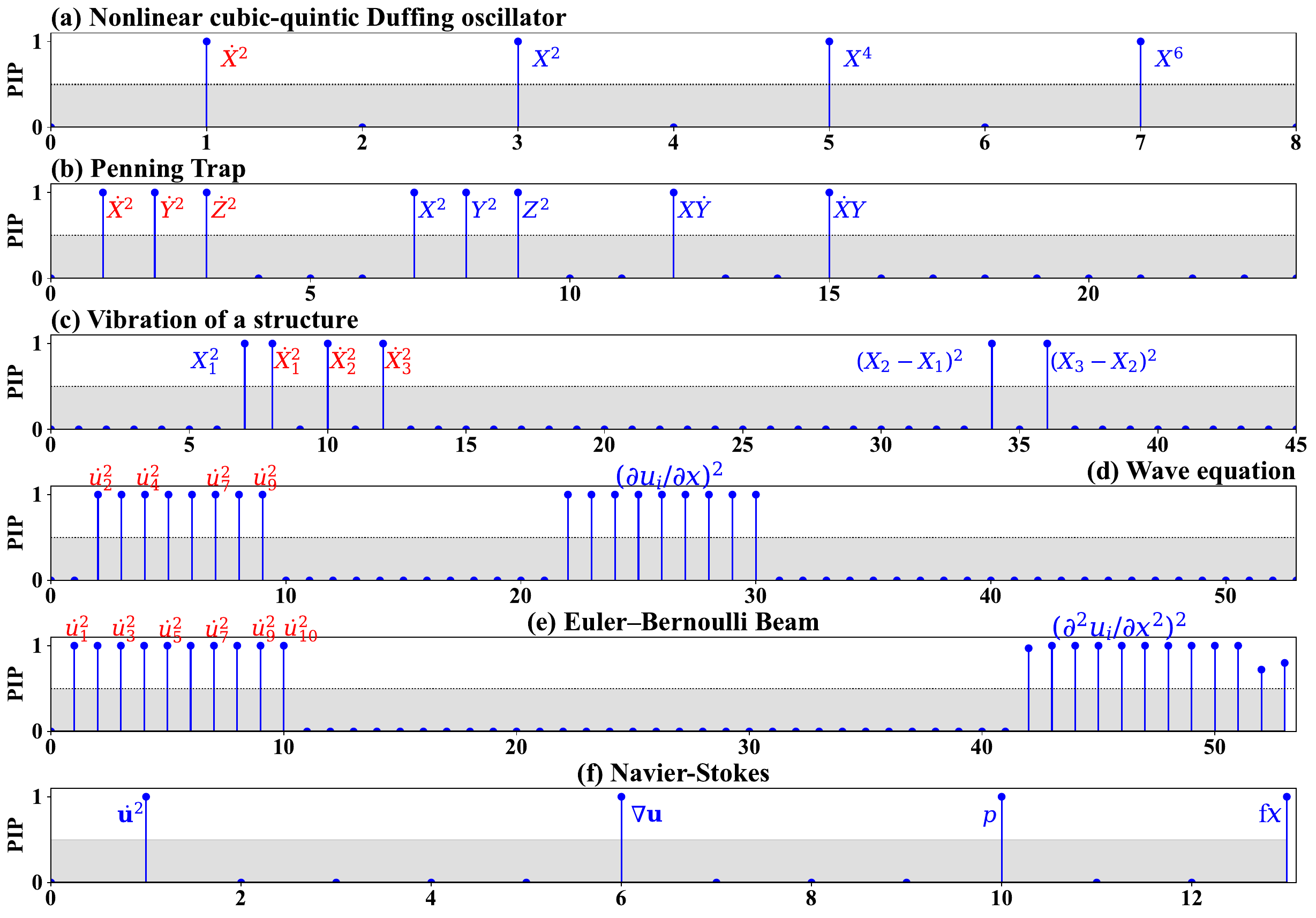}    
    \caption{{Posterior inclusion probability (PIP) of the basis functions}. The PIP values are estimated after discarding the burn-in samples of MCMC iterations. The basis functions with PIP$>0.5$ are included in the final model. The black dotted line denotes PIP=$0.5$. The Lagrangian is identified at each DOF in the discrete systems in (a), (b), and (c). The Lagrangian is identified at each discretized location in continuous systems (d) and (e). For the Navier-Stokes equation, the Lagrangian is identified at the global level by reshaping the data matrix into vectors.}
    \label{fig:stem_basis}
\end{figure}
\subsection{Example 1: Nonlinear cubic-quintic Duffing oscillator}\label{example_2}
\begin{figure}[ht!]
    \centering     
    \includegraphics[width=0.3\textwidth]{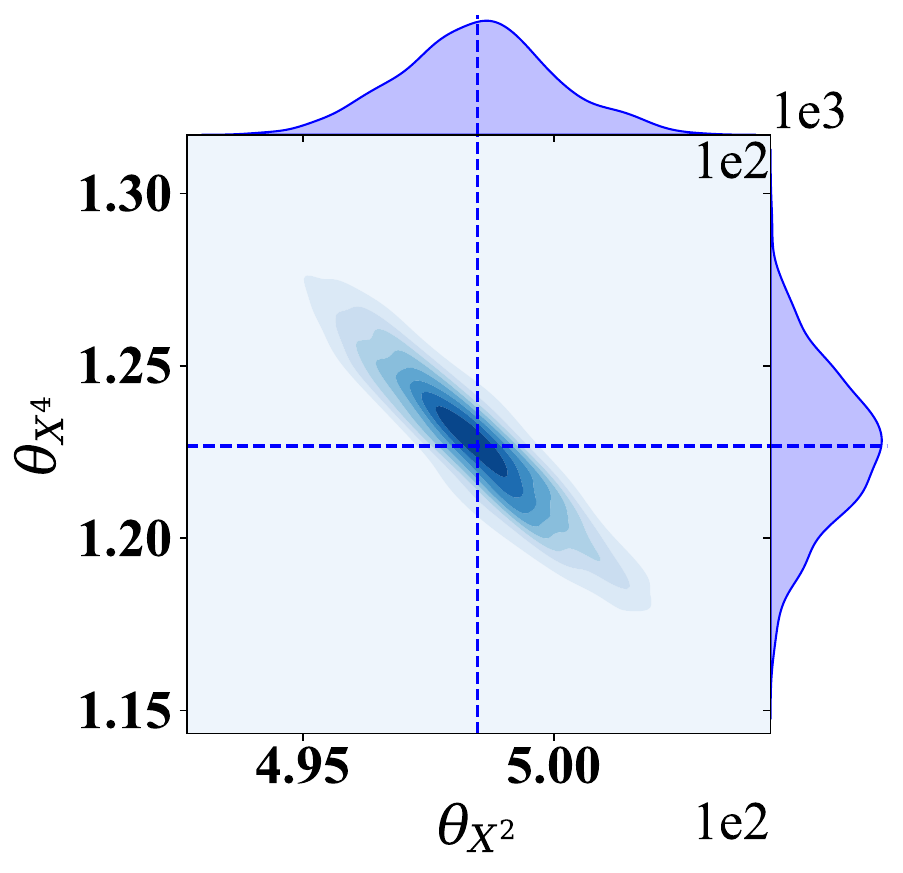}
    \includegraphics[width=0.3\textwidth]{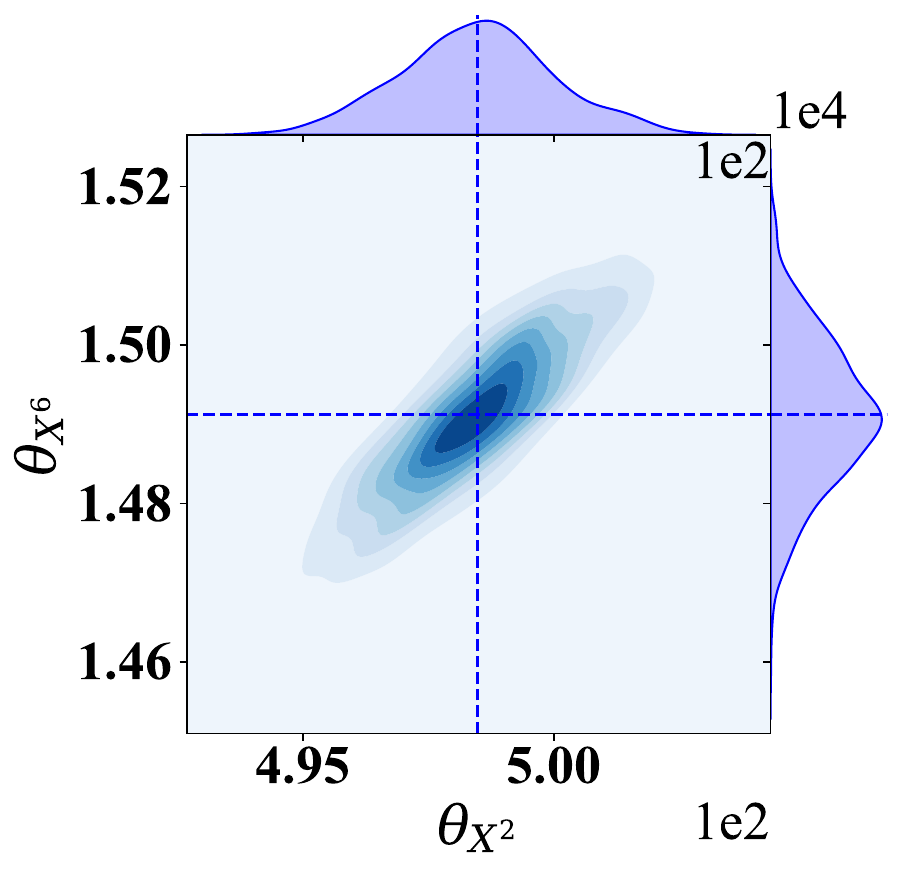}
    \includegraphics[width=0.3\textwidth]{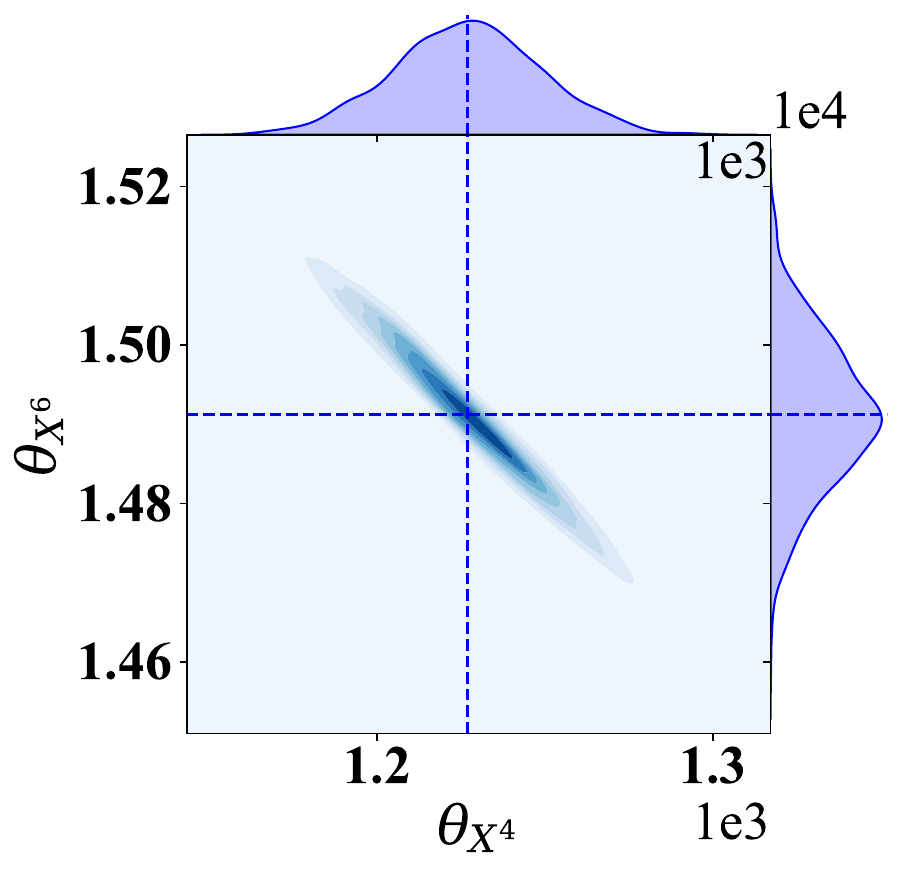}
    \caption{{Nonlinear cubic-quintic Duffing oscillator}: The pairwise joint posterior probability distributions of the coefficients $\theta_{X^2}$, $\theta_{X^4}$, and $\theta_{X^6}$ associated with the basis functions ${X^2}$, ${X^4}$, and ${X^6}$, respectively. The blue dash--lines indicate the expected value of the corresponding parameters. A total of 4000 MCMC iterations are used, and the first 1500 samples are discarded as burn-in.}
    \label{fig:param_nonlinear}
\end{figure}

In the first example, we aim to learn the exact analytical form of the Lagrangian of a nonlinear cubic-quintic Duffing oscillator, for which the training data is generated using the following governing equation of motion,
\begin{equation}
    \ddot{X}(t) + \alpha {X}(t) + \beta {X}^{3}(t) + \gamma {X}^{5}(t) = 0; \; X(t=0)=X_0, \; t \in [0,T],
\end{equation}
where $\alpha \in \mathbb{R}$, $\beta \in \mathbb{R}$, and $\gamma \in \mathbb{R}$ are the real system constant parameters. For the simulation of the training data, the system parameters are taken as $\alpha$=1000, $\beta$=5000, and $\gamma$=90000. The initial displacement and velocity of the nonlinear oscillator are considered as $\{X, \dot{X}\}(0)$=\{0.35,0\}. The training data is generated for $T$=0.5s with a sampling frequency of 2000Hz. Upon performing the sparse Bayesian regression, only those basis functions with a PIP value of $\ge$0.5 are selected to constitute the final Lagrangian model.
\textbf{Results}: The PIP values of the basis functions obtained after discarding burn-in samples are shown in Fig. \ref{fig:stem_basis}. The design matrix $\mathbf{L}\in \mathbb{R}^{N\times 9}$ contains 9 basis functions and correspondingly 9 sparse coefficients. It is evident in the PIP plot that the PIP values for basis functions other than the true ones, i.e., $\dot{X}^2$, $X^2$, $X^4$, and $X^6$ are almost zero, indicating the parsimonious nature of the discovered model. The pairwise joint posterior probability distributions of the associated parameters of the basses $X^2$, $X^4$, and $X^6$ are provided in Fig. \ref{fig:param_nonlinear}. The final Lagrangian form and the relative $L^2$ norm of the model parameter identification error are provided in Table \ref{tab:ident_param}, where it is evident that the rediscovered Lagrangian of the nonlinear oscillator almost accurately approximates the true Lagrangian density.

\begin{table}[!ht]
    \centering
    \caption{Summary of the results of the Lagrangian discovery of example problems. The relative $L^2$ error is estimated between the parameters of the actual and identified Lagrangian densities as $\|\bm{\beta}-\bm{\beta}^{*}\|/\|\bm{\beta}\|$, where $\bm{\beta}$ and $\bm{\beta}^{*}$ represents the parameters of the actual and identified Lagrangian density, and $\|\cdot\|$ denote the usual $L^2$ norm. Here $\mathcal{L}$ and $\mathcal{L}^{*}$ denotes the actual and discovered Lagrangian density.}
    \begin{tabular}{lc}
        \toprule
        Systems/Lagrangian & Relative $L^2$ error \\
        \midrule
        Nonlinear cubic--quintic Duffing oscillator & \\ \cmidrule(r){1-2}
        $\mathcal{L}$=$\frac{1}{2}\dot{X}^2 - 500X^2 - 1250X^4 - 15000X^6$ & -- \\
        $\mathcal{L}^{*}$=$\frac{1}{2}\dot{X}^2 - {498.46}{\scriptstyle\textcolor{blue}{\pm 1.50}}X^2 - {1226.76}{\scriptstyle\textcolor{blue}{\pm 21.48}}X^4 - {14912.12}{\scriptstyle\textcolor{blue}{\pm 92.11}}X^6$ & $0.6037{\scriptstyle \textcolor{blue}{\pm 0.03}}$ \\
        \midrule
        Penning Trap & \\ \cmidrule(r){1-2}
        $\mathcal{L}$=$\frac{1}{2} \left( \dot{X}^2 + \dot{Y}^2+ \dot{Z}^2 \right) + 25\left( X^2 + Y^2 - 2Z^2 \right) + 50\left( X\dot{Y} + \dot{X}Y \right)$ & -- \\
        $\mathcal{L}^{*}$=$\frac{1}{2} \left( \dot{X}^2 + \dot{Y}^2+ \dot{Z}^2 \right) + 24.99{\scriptstyle\textcolor{blue}{\pm 0.01}}\left( X^2 + Y^2 - 2Z^2 \right) + 49.95{\scriptstyle\textcolor{blue}{\pm 0.66}}\left( X\dot{Y} + \dot{X}Y \right)$ & $0.0912{\scriptstyle\textcolor{blue}{\pm 0.01}}$ \\ 
        \midrule
        Vibration of structure & \\ \cmidrule(r){1-2}
        $\mathcal{L}$=$\frac{1}{2} \sum_{k=1}^{3} \dot{X}_k^2 - 2500\left( X_1^2 + (X_2-X_1)^2 + (X_3-X_2)^2 \right)$ & -- \\
        $\mathcal{L}^{*}$=$\frac{1}{2} \sum_{k=1}^{3} \dot{X}_k^2 -  {2497.68}{\scriptstyle\textcolor{blue}{\pm 0.18}} X_1^2 - {2497.67}{\scriptstyle\textcolor{blue}{\pm 0.23}}(X_2-X_1)^2 - {2497.14}{\scriptstyle\textcolor{blue}{\pm 0.41}}(X_3-X_2)^2 $ & $0.1006{\scriptstyle \textcolor{blue}{\pm 0.09}}$ \\ 
        \midrule
        Vibration of string & \\ \cmidrule(r){1-2}
        $\mathcal{L}$=$\frac{1}{2}\sum_{i} \dot{u}_i^2 - \sum_i 10 (\partial_x u_{i})^2$ & -- \\
        $\mathcal{L}^{*}$=$\frac{1}{2}\sum_{i} \dot{u}_i^2 - \sum_i {9.98}{\scriptstyle \textcolor{blue}{\pm 0.05}} (\partial_x u_{i})^2$ & $0.19{\scriptstyle \textcolor{blue}{\pm 0.31}}$ \\
        \midrule
        Vibration of Euler–Bernoulli Beam & \\ \cmidrule(r){1-2}
        $\mathcal{L}$=$\frac{1}{2} \sum_{i} \dot{u}_i^2 - 2.1231 \sum_i (\partial_{xx} u_{i})^2$ & -- \\
        $\mathcal{L}^{*}$=$\frac{1}{2} \sum_{i} \dot{u}_i^2 - {2.1168}{\scriptstyle \textcolor{blue}{\pm 0.01}} \sum_i (\partial_{xx} u_{i})^2$ & $0.2980{\scriptstyle \textcolor{blue}{\pm 0.28}}$ \\
        \midrule
        Navier-Stokes equation & \\ \cmidrule(r){1-2}
        $\mathcal{L}$=$\frac{1}{2} \dot{\bm{u}}^2 -0.00125p + 0.0001\nabla \bm{u} + f\cdot\bm{I}x $ & -- \\
        $\mathcal{L}^{*}$=$\frac{1}{2} \dot{\bm{u}}^2 -\underset{{\scriptstyle \textcolor{blue}{\pm 2.21\times 10^{-12}}}}{0.00112}p + [\underset{{\scriptstyle \textcolor{blue}{\pm 7.16\times 10^{-12}}}}{0.0002},\underset{{\scriptstyle \textcolor{blue}{\pm 1.28\times 10^{-13}}}}{0.0001}]^{\top} \cdot \nabla \bm{u} + \underset{{\scriptstyle \textcolor{blue}{\pm 4.82\times10^{-7}}}}{0.9841}f\cdot\bm{I}x $ & $0.0159{\scriptstyle \textcolor{blue}{\pm 1.31\times 10^{-6}}}$ \\
        \bottomrule
    \end{tabular}
    \label{tab:ident_param}
\end{table}

\subsection{Example 2: Penning Trap}\label{example_penning}
\begin{figure}[ht!]
    \centering     
    \includegraphics[width=0.3\textwidth]{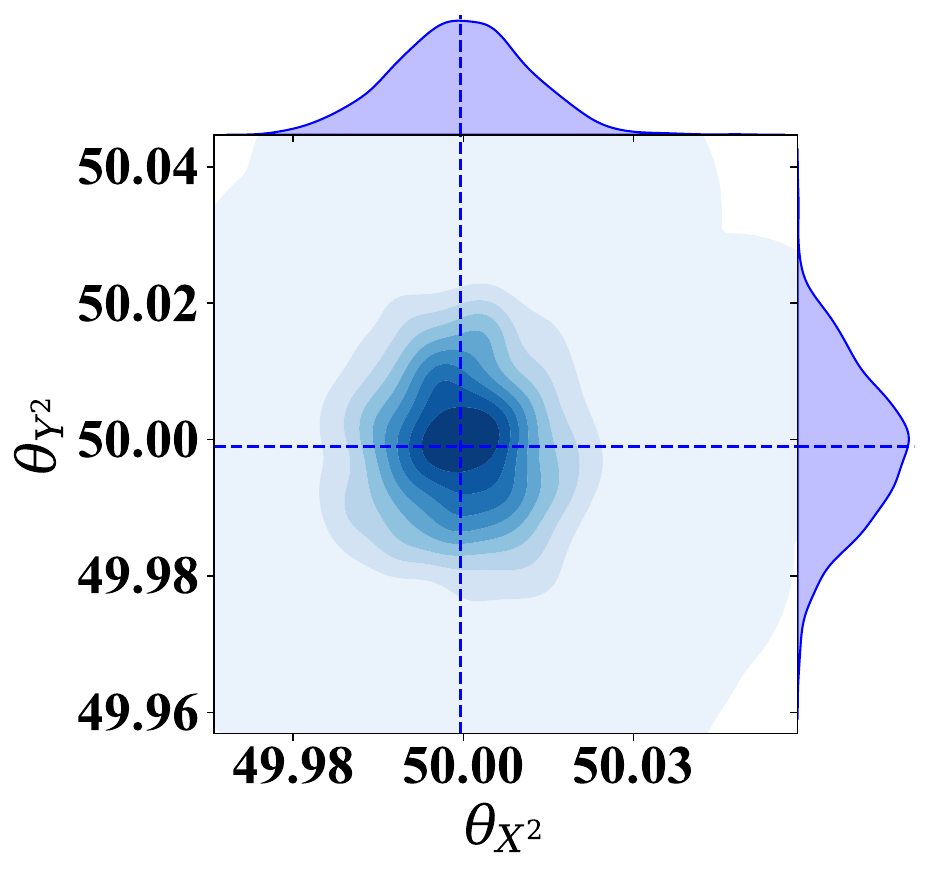}
    \includegraphics[width=0.3\textwidth]{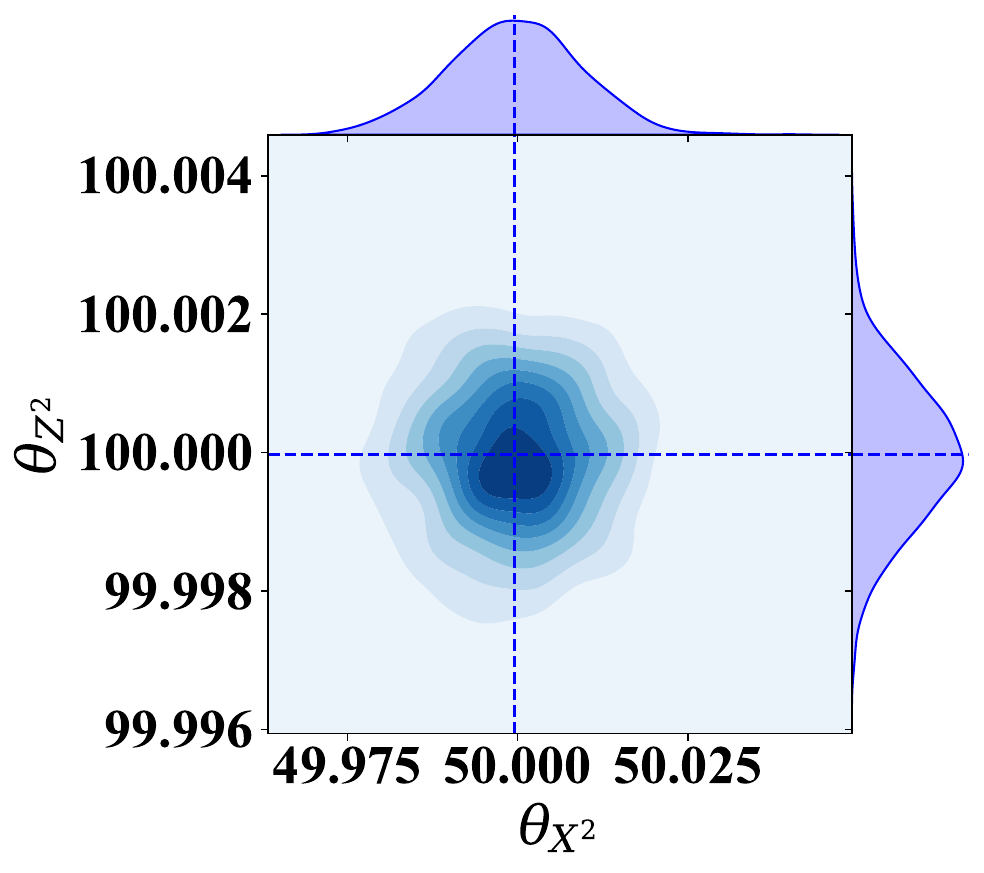}
    \includegraphics[width=0.3\textwidth]{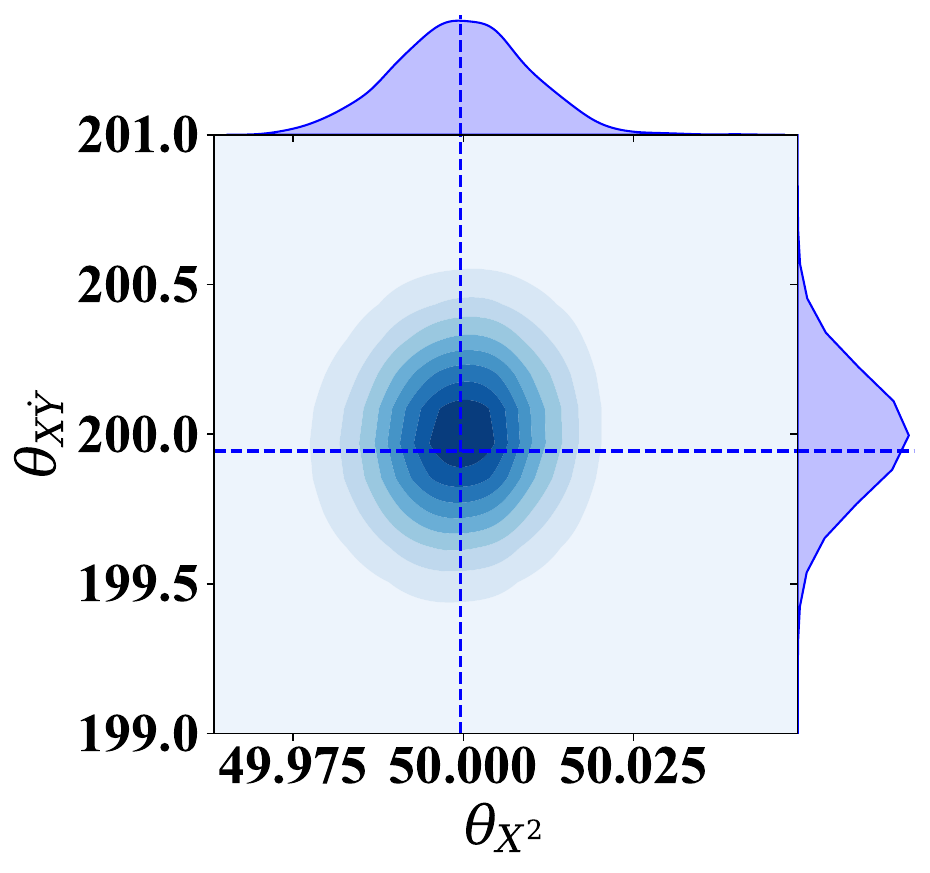}
    \includegraphics[width=0.3\textwidth]{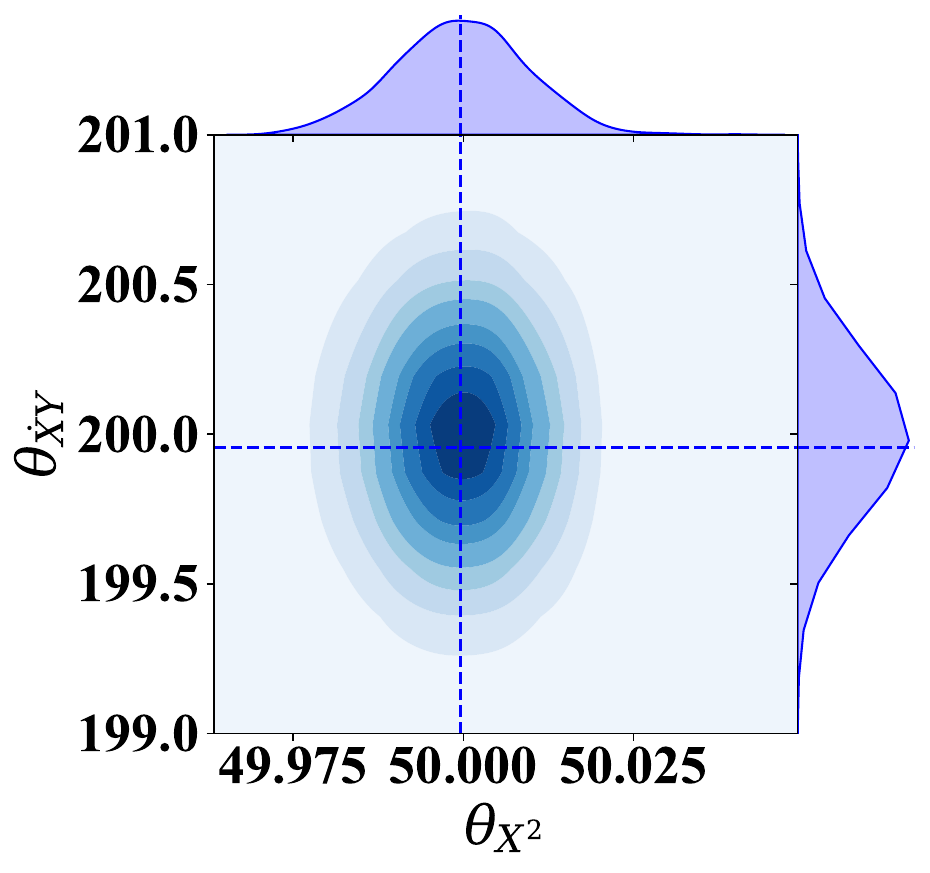}
    \includegraphics[width=0.3\textwidth]{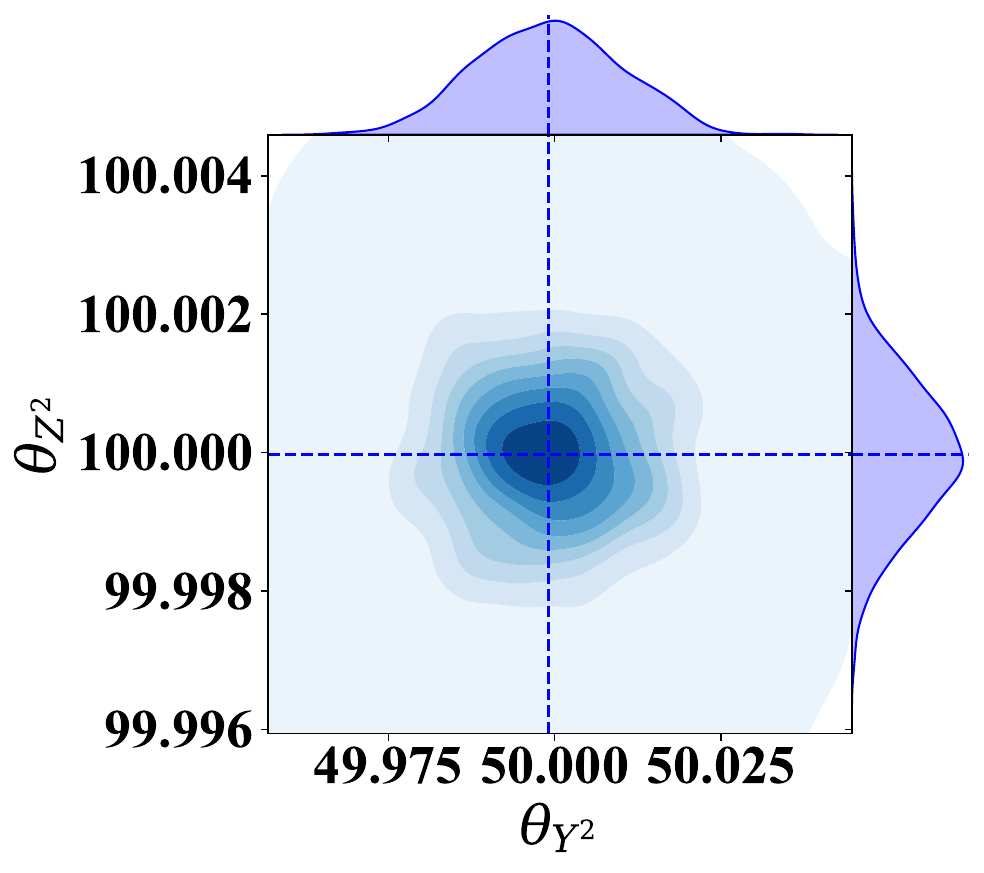}
    \includegraphics[width=0.3\textwidth]{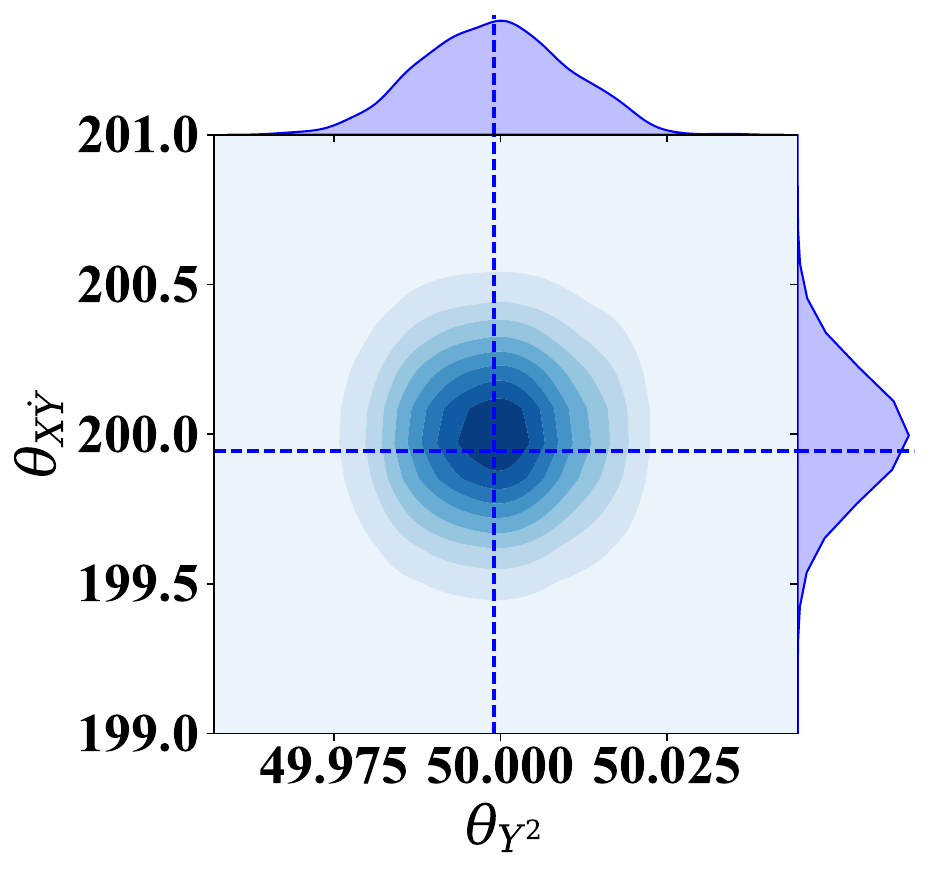}
    \includegraphics[width=0.3\textwidth]{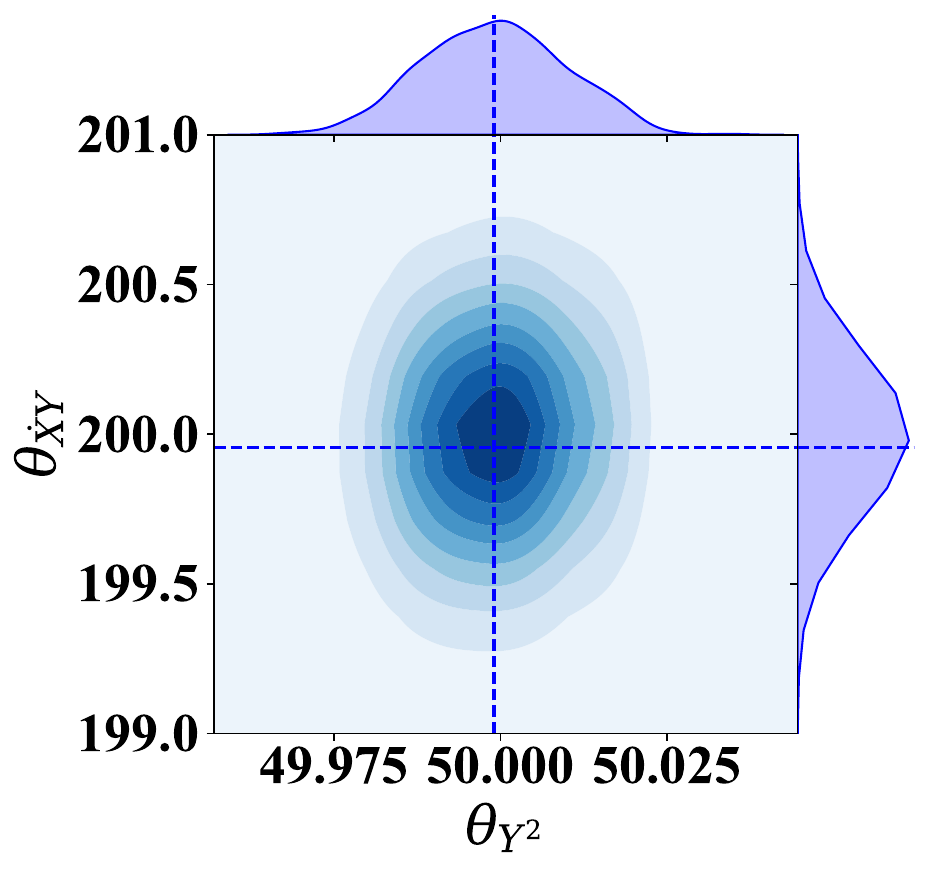}
    \includegraphics[width=0.3\textwidth]{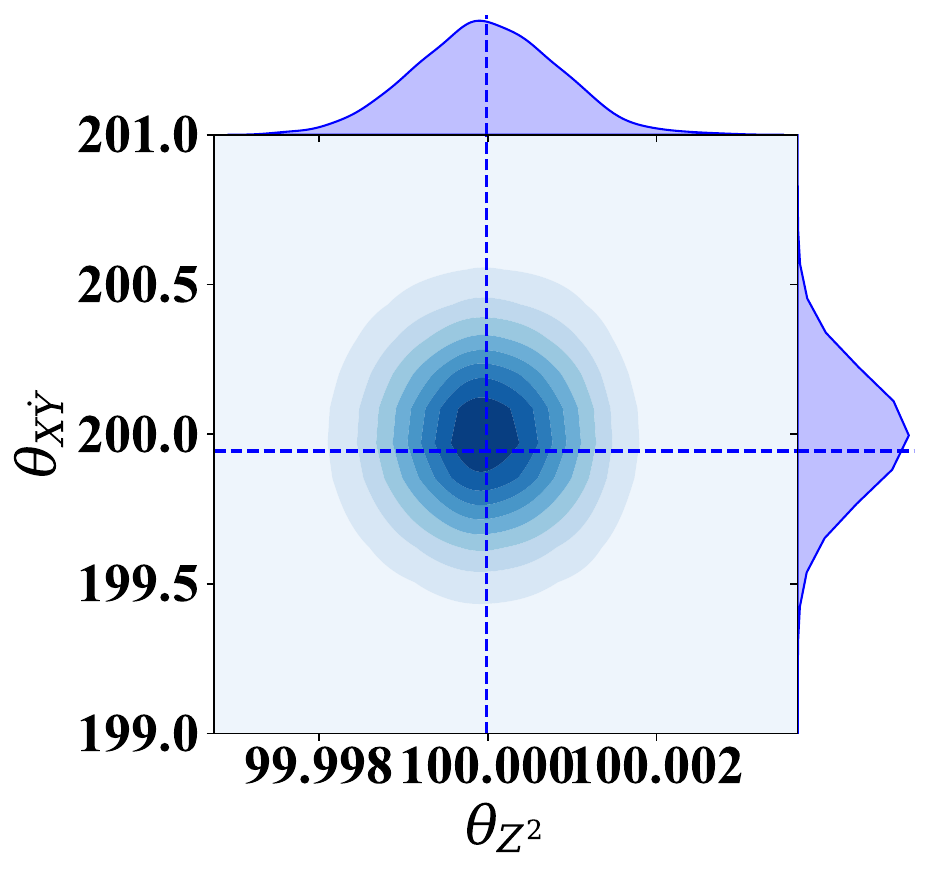}
    \includegraphics[width=0.3\textwidth]{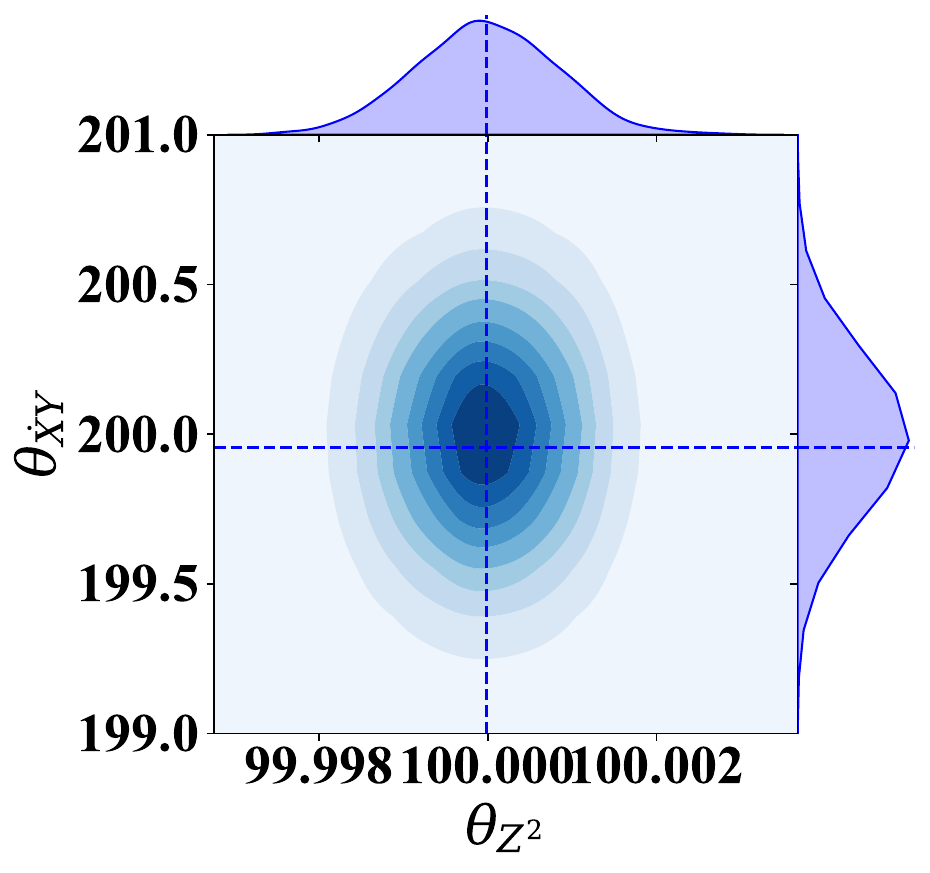}
    \caption{{Penning Trap}: The pairwise joint posterior probability distributions of the coefficients $\theta_{X^2}$, $\theta_{Y^2}$, $\theta_{Z^2}$, $\theta_{X\dot{Y}}$, and $\theta_{\dot{X}Y}$ associated with the basis functions ${X^2}$, ${Y^2}$, $Z^2$, $X\dot{Y}$ and ${\dot{X}Y}$, respectively. The blue dash--lines indicate the expected value of the corresponding parameters. A total of 3000 MCMC iterations are used, and the first 1000 samples are discarded as burn-in.}
    \label{fig:param_trap}
\end{figure}

In the second example, we consider the drift of a charged particle in a radial electric field. A Penning trap is a versatile tool used for storing charged particles and for precision measurements of properties of subatomic particles, e.g., mass and fission yields. We aim to discover the Lagrangian that governs the motion of the charged particle inside the Penning trap. For simulating the training data, describing the drift of a charged particle in the radial electric field of the penning trap, we follow the following equations,
\begin{equation}
    \begin{aligned}
        m \ddot{X}(t) &= q_e \left( B\dot{Y}(t) + \frac{1}{2}kX(t) \right), \\
        m \ddot{Y}(t) &= q_e \left( -B\dot{X}(t) + \frac{1}{2}kY(t) \right), \\
        m \ddot{Z}(t) &= - q_e k Z(t) ,
    \end{aligned}
\end{equation}
where $m$ is the charged particle's mass, and $q_e$ is the particle's electric charge. For convenience, we introduce the cyclotron frequency $\omega_c = q_e B/m$ and the axial frequency $\omega_a = \sqrt{q_e k/m}$. For generating synthetic data, we consider $\omega_c=100$Hz, $\omega_a=10$Hz, and $\bm{X}(0)=\{10^{-3},0,10^{-3},0,10^{-2},0\}$. The training data is simulated for $T=0.3$s at a sampling frequency of 10000Hz. A PIP value of 0.5 is considered to include the identified basis functions in the final Lagrangian model. 
\textbf{Results}: The design matrix $\mathbf{L}\in \mathbb{R}^{N \times 25}$ for this example contains 25 basis functions and 25 regression coefficients. The PIP values of the basis functions after discarding the burn-in MCMC samples are provided in Fig. \ref{fig:stem_basis}. The identified basis functions in the discovered Lagrangian model for the charged particle are found to be $\dot{X}^2$, $\dot{Y}^2$, $\dot{Z}^2$, $X^2$, $Y^2$, $Z^2$, $X\dot{Y}$, and $Y\dot{X}$, indicating that only 8 basis function are selected in the final discovered model with a probability more than 0.5. The pairwise joint posterior distributions of the identified parameters are provided in Fig. \ref{fig:param_trap}. For an overall comparison, the final Lagrangian is provided in Table \ref{tab:ident_param}, where it is evident that the proposed algorithm approximates the true Lagrangian density very accurately.

\subsection{Example 3: Vibration of a 3DOF structure}\label{example_4}
\begin{figure}[t!]
    \centering
    \includegraphics[width=0.3\textwidth]{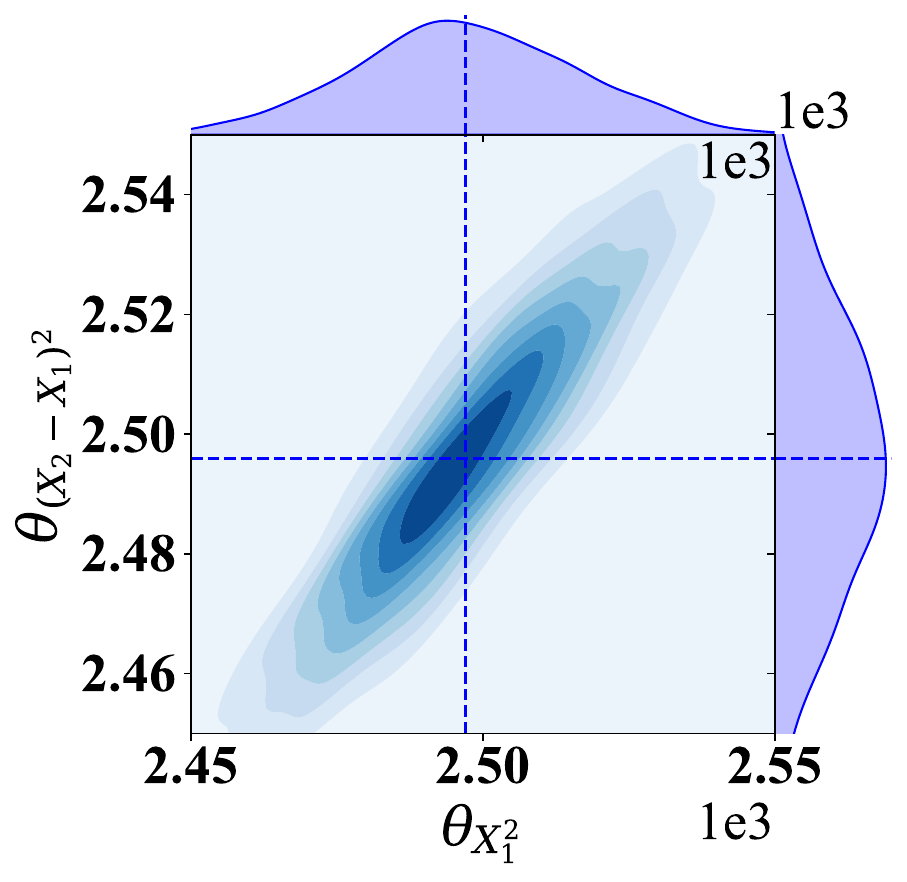}
    \includegraphics[width=0.3\textwidth]{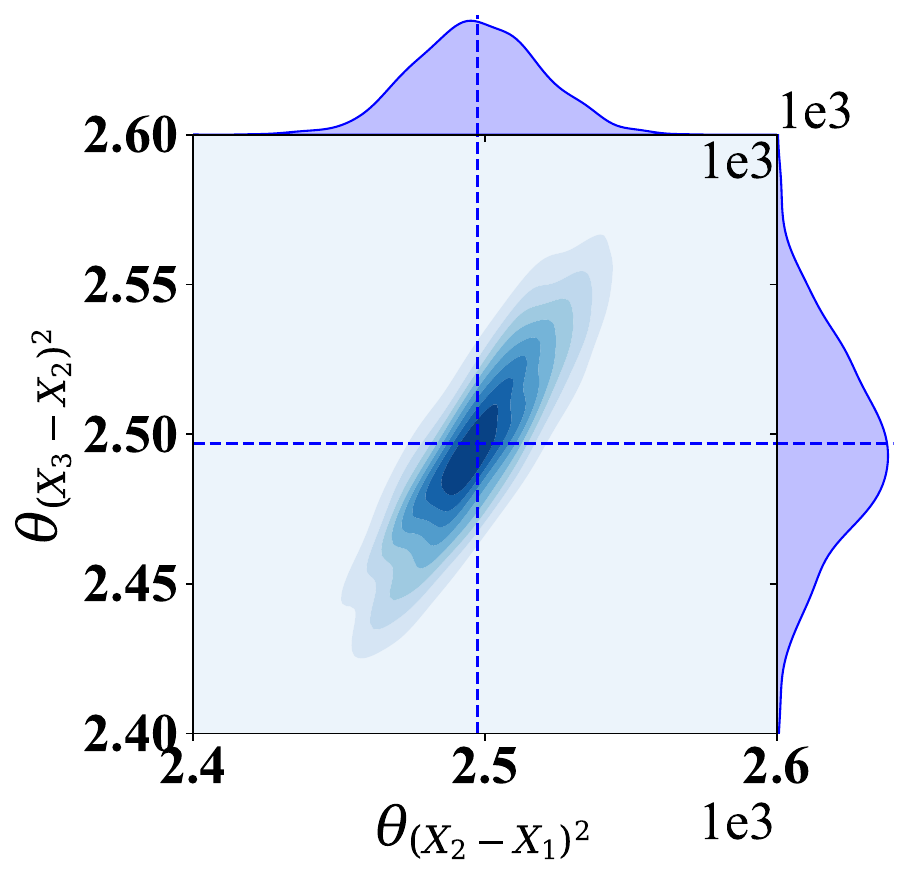}
    \includegraphics[width=0.3\textwidth]{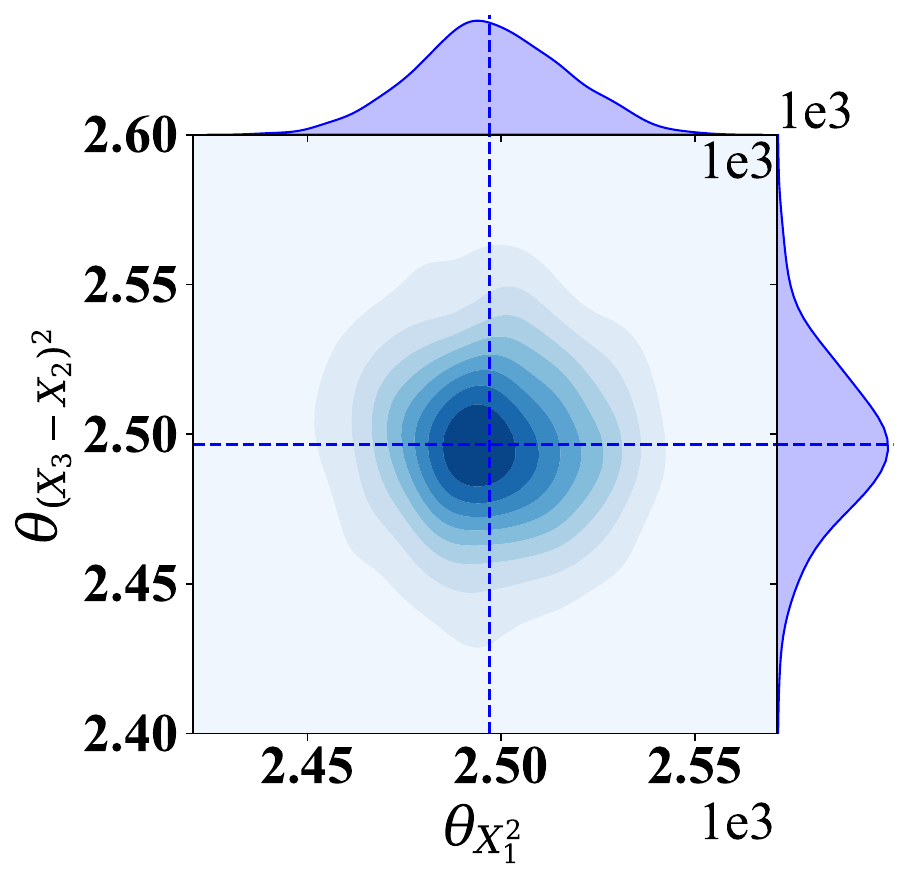}
    \caption{{Undamped MDOF vibration}: Pairwise joint posterior probability distributions of the coefficients $\theta_{{X}_1^2}$, $\theta_{(X_2-X_1)^2}$, and $\theta_{(X_3-X_2)^2}$ associated with the basses ${X}_1^2$, $(X_2-X_1)^2$, and $(X_3-X_2)^2$, respectively. The blue lines indicate the expected value of the identified parameters. A total of 4000 MCMC iterations are used, and the first 1500 samples are discarded as burn-in.}
    \label{fig:param_3dof}
\end{figure}

In this example, we want to discover the Lagrangian associated with the undamped vibration of a 3DOF structural system. The structure is idealized as a mass-spring-dashpot system, for which the training data is simulated using the following equation,
\begin{equation}\label{eq:mdof}
    \mathbf{M} \ddot{\bm{X}}(t) + \mathbf{K} \bm{X}(t) = 0, \; \bm{X}(t=t_0)=\bm{X}_0, \; t \in [0,T],
\end{equation}
where the mass matrix $\mathbf{M} \in \mathbb{R}^{3 \times 3}$ and the stiffness matrix $\mathbf{K} \in \mathbb{R}^{3 \times 3}$ are given as,
\begin{equation}
    \mathbf{M} = \left[ { \begin{array}{ccc}
         m_1 & 0 & 0 \\
         0 & m_2 & 0 \\
         0 & 0 & m_3
    \end{array} } \right], \; \text{and,} \;
    \mathbf{K} = \left[ { \begin{array}{ccc}
         k_1+k_2 & -k_2 & 0 \\
         -k_2 & k_2+k_3 & -k_3 \\
         0 & -k_3 & k_3
    \end{array} } \right].
\end{equation}
The following parameters are used for training data simulation: $m_i$=1kg, $k_i$=5000N/m$^2$ for $i=1,2,3$. The initial conditions are taken as $\bm{X} = \{1,0,2,0,3,0\}$, and the training data is generated for $T=1$s using a sampling frequency of 1000Hz. 
\textbf{Results}: For this example, the design matrix $\mathbf{L}\in \mathbb{R}^{N \times 46}$ is constructed using 46 basis functions. The PIP values of the basis functions after discarding the burn-in MCMC samples are shown in Fig. \ref{fig:stem_basis}. We observe that only the basis functions that are present in the actual model, i.e., $\dot{X}_1^2$, $\dot{X}_2^2$, $\dot{X}_3^2$, $X_1^2$, $(X_2-X_1)^2$, and $(X_3-X_2)^2$ have a PIP value greater than 0.5, indicating the sparse nature of the parameter vector $\bm{\theta} \in \mathbf{R}^{46}$. The pairwise joint posterior probability distributions of the parameters of $X_1^2$, $(X_2-X_1)^2$, and $(X_3-X_2)^2$ are shown in Fig. \ref{fig:param_3dof}. The analytical form of the learned Lagrangian is compared against the actual in Table \ref{tab:ident_param}, where we observe that the proposed framework accurately approximates the system parameters.

\subsection{Example 4: Vibration of string}\label{example_6}
\begin{figure}[t!]
    \centering
    \includegraphics[width=0.3\textwidth]{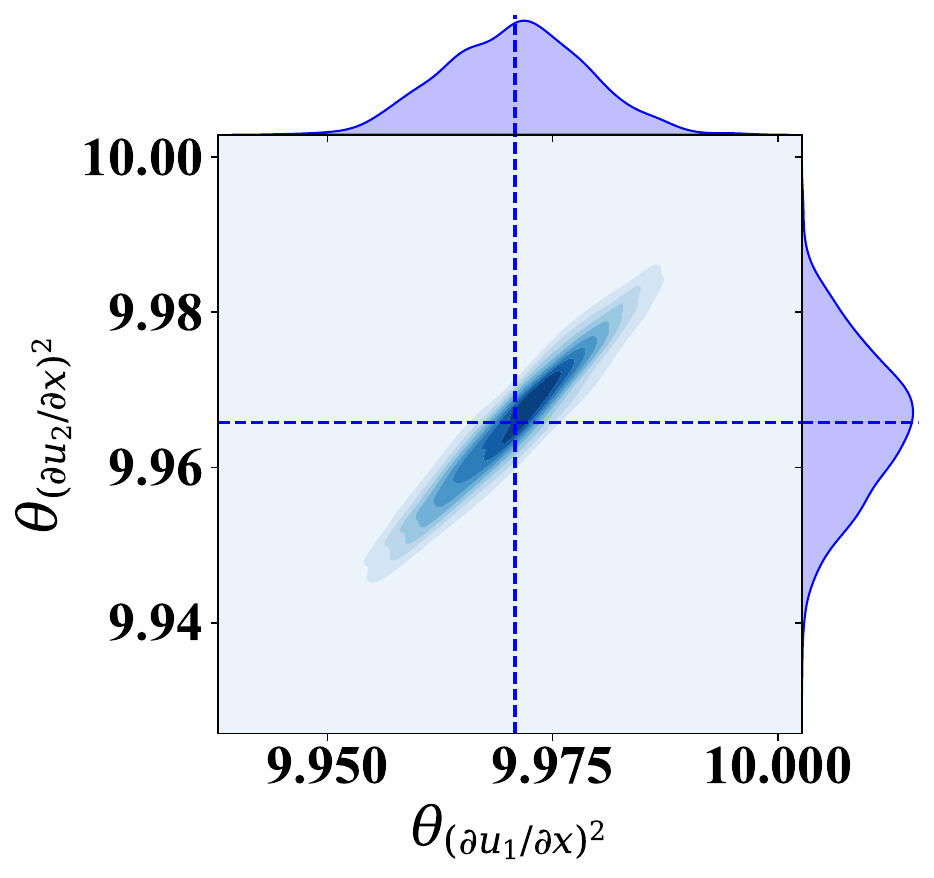}
    \includegraphics[width=0.3\textwidth]{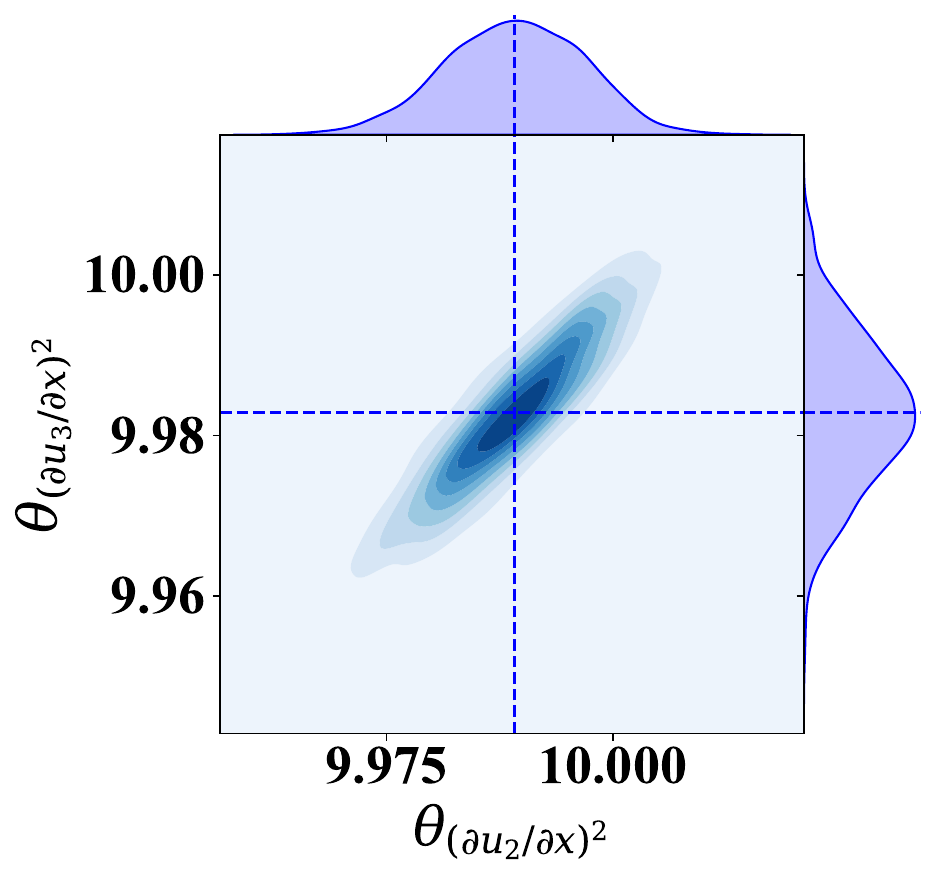}
    \includegraphics[width=0.3\textwidth]{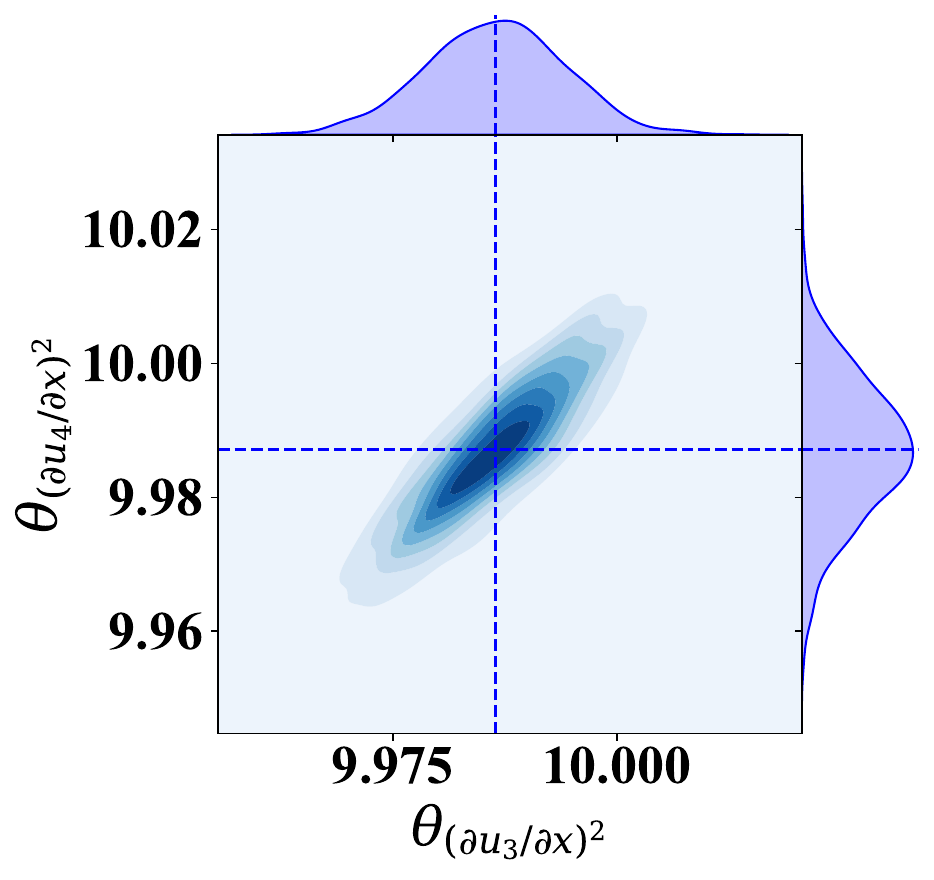}
    \includegraphics[width=0.3\textwidth]{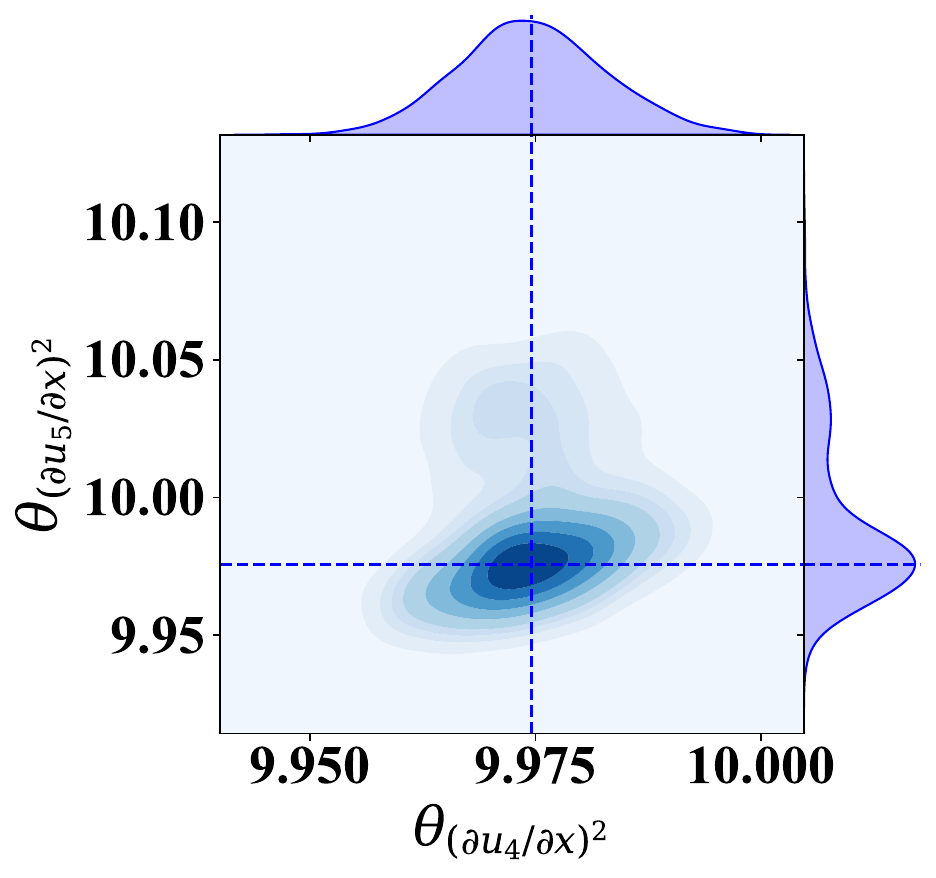}
    \includegraphics[width=0.3\textwidth]{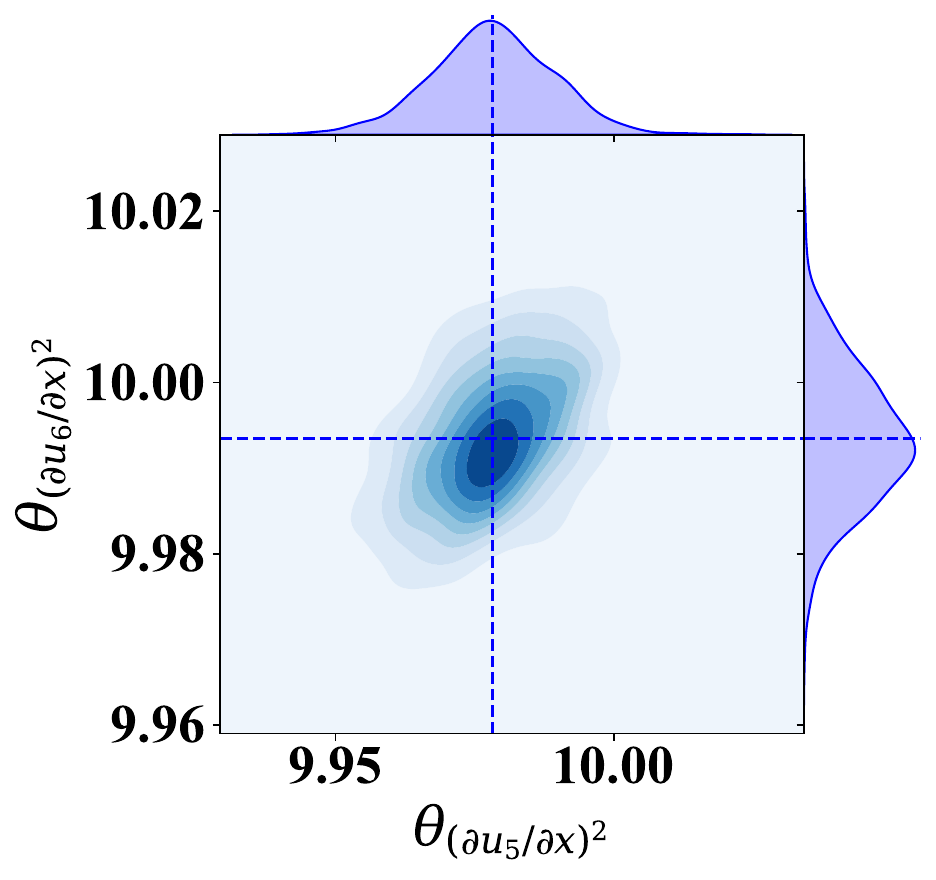}
    \includegraphics[width=0.3\textwidth]{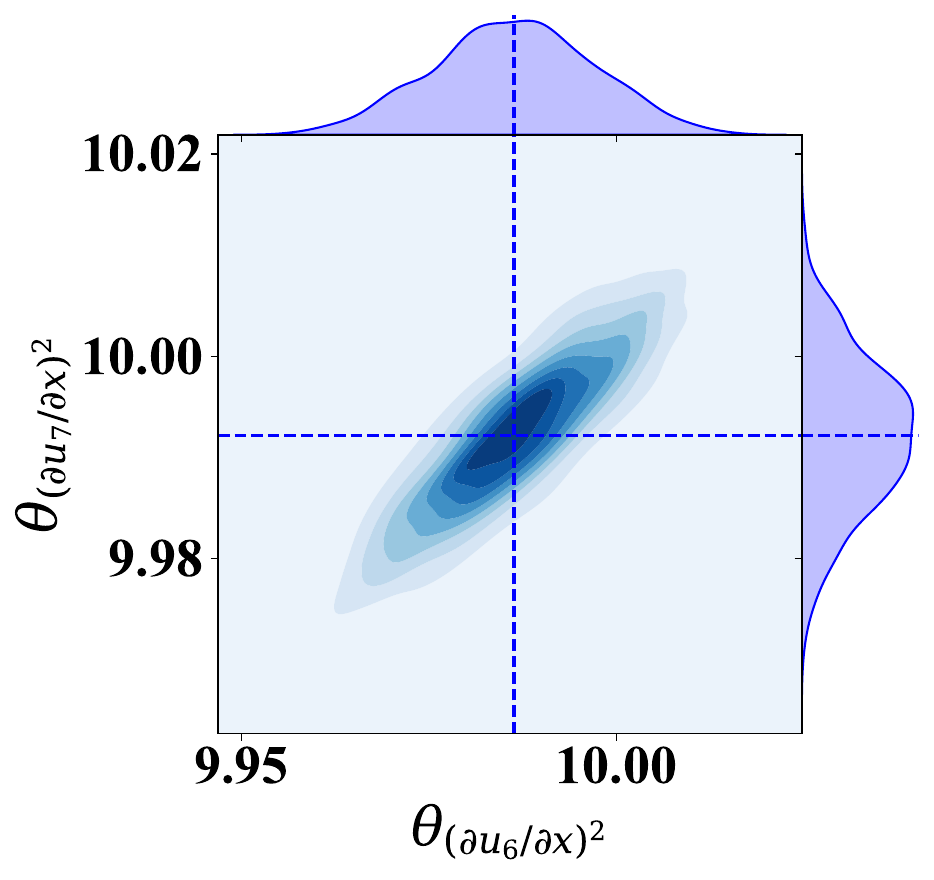}
    \includegraphics[width=0.3\textwidth]{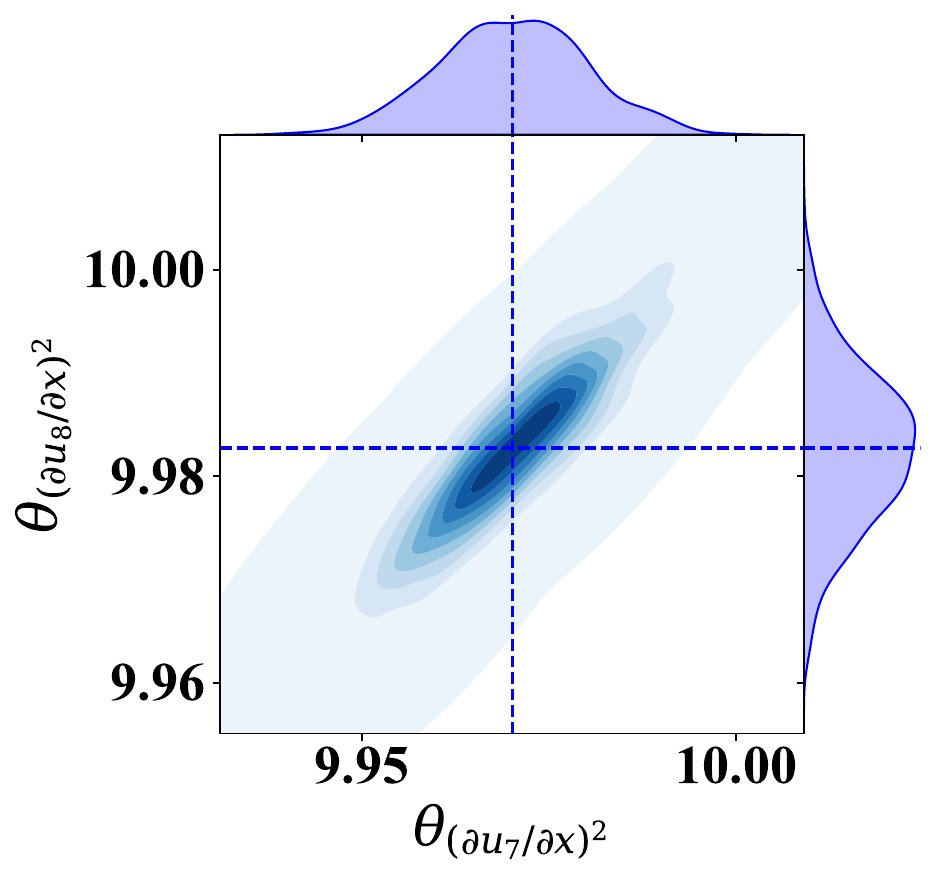}
    \includegraphics[width=0.3\textwidth]{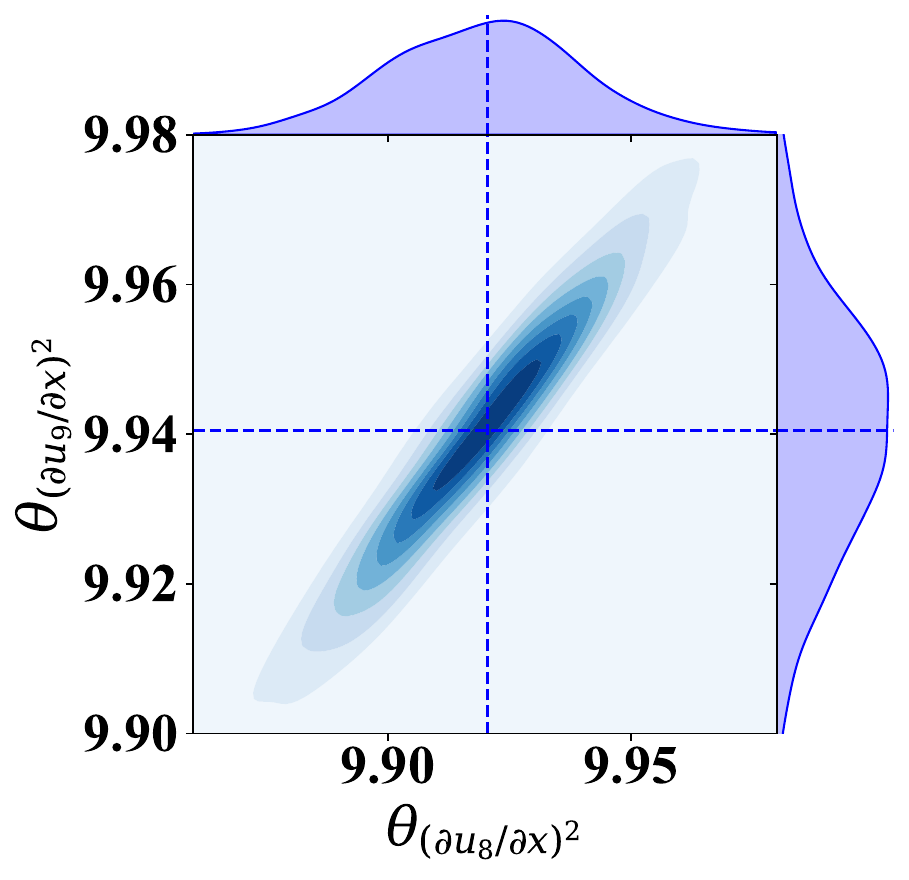}
    \caption{{Elastic Transversal waves in solid}: Joint posterior probability distributions of the coefficients of the basis functions $\theta_{(\partial u_{i}/ \partial x)^2}$. The blue lines indicate the expected value of the identified parameters. A total of 5000 MCMC iterations are used, and the first 2000 samples are discarded as burn-in.}
    \label{fig:param_string}
\end{figure}

Here, we focus on discovering the Lagrangian of a continuous system, i.e., a system having spatio-temporal activity. In particular, we consider the propagation of elastic transversal waves in strings. The following partial differential equation is used to generate the training data for the propagation of the transversal waves in a string,
\begin{equation}
    \begin{aligned}
        & \partial_{tt} u(X,t) - c^2 \partial_{xx} u(X,t) = 0, \;\; x \in [0,1], \; t\in [0,T] \\
        & u(0,t)=0, u(1,t)=0, u_0(x,0)={\rm{cos}}(2\pi L),
    \end{aligned}
\end{equation}
where $c=\sqrt{\mu / \rho}$ is the speed of elastic transversal wave with $\mu$ being the shearing modulus, and $\rho$ being the mass density of the material. 
For training data simulation, the coefficient $c$ and the length $L$ are taken as 10m/s and 1m. We discretize $L$ into a 10-dimensional spatial grid using $\Delta x$=0.1m, and at each node of the grids, the system response is simulated for $T$=1s with a time step $\Delta t$=0.001s. The Lagrangian density of the system is discovered at each discretized location. 
\textbf{Results}: The design matrix $\mathbf{L}$ has $\mathbb{R}^{N \times 54}$ has a total of 54 basis functions and a 54-dimensional parameter vector $\bm{\theta} \in \mathbf{R}^{54}$. The PIP values of the basis functions after discarding the burn-in MCMC samples are shown in Fig. \ref{fig:stem_basis}. Evidently, the proposed framework correctly identified the exact basis functions $(\partial u_i/ \partial t)^2$ and $(\partial u_i/ \partial x)^2$ for $i=1,\ldots,9$. The probability distributions of the basis coefficients are illustrated in Fig. \ref{fig:param_string}, where we observe that the coefficient values of the identified basis functions at all the spatial locations approximate the ground truth values very accurately. The Lagrangian of the complete system, which is obtained by summing the Lagrangian over all the spatial locations, is given in Table \ref{tab:ident_param}.  

\subsection{Example 5: Vibration of Euler-Bernoulli beam}\label{example_7}
Here, we consider the flexion vibration of an Euler-Bernoulli beam, where we aim to discover the Lagrangian describing the dynamics behind the vibration of this beam. We neglect the effect of gravity and assume that the beam has a homogeneous and uniform section with linear mass $\mu$ and length $L$. The training data for the Euler-Bernoulli beam is generated using the following differential equation,
\begin{equation}
    \begin{aligned}
        & \partial_{tt} u(X,t) = c \partial_{xxxx} u(X,t), \;\; x\in [0,1], \; t \in [0,T] \\
        & u(x=0, t) = 0, \; \partial_x u(x=0, t) = 0, \; \partial_{xx} u(x=L, t) = 0, \; \partial_{xxx} u(x=L, t) = 0, \\
        & u(x, t=0) = (\operatorname{cosh}\varphi x - \operatorname{cos}\varphi x) + \frac{(\operatorname{cos}\varphi L + \operatorname{cosh}\varphi L)}{(\operatorname{sin}\varphi L + \operatorname{sinh}\varphi L)} (\operatorname{sin}\varphi x - \operatorname{sinh}\varphi x),
    \end{aligned}
\end{equation}
where $c= {EI}/{\mu}$ with $E$ being the modulus of elasticity of the material, and $I$ is the moment-of-inertia. For synthetic data generation, $c$ and $\varphi$ in this study are taken as 2.12m$^2$/s ($\rho$=7850kg/m$^3$, $E$=2$\times 10^{11}$N/m$^2$, $L$=1, $b$=0.02m, $d$=0.001m) and $3.5\pi$.
The training data is obtained by vibrating the beam's fourth natural frequency for $T$=0.1s using $\Delta t$=0.0001s. The Spatial discretization of $L$ is performed using N=10 segments of length $\Delta x$=0.1m. The Lagrangian of the system is given in Table \ref{tab:ident_param}, where $u_{i}$ denotes the deformation at the center of mass of the $i^{th}$ segment from the equilibrium point. Similar to the previous example, the Lagrangian density is discovered at the discretized locations. 
\textbf{Results}: The regression model for this example has 54 sparse coefficients associated with 54 basis functions. The basis functions that have a PIP value higher than 0.5 are portrayed in Fig. \ref{fig:stem_basis}, where we observe that the proposed sparse Bayesian framework is able to correctly identify the exact basis functions $(\partial u_i / \partial t)^2$ and $(\partial u_i^2 / \partial x^2)^2$. The final form of the Lagrangian, along with model uncertainties, are provided in Table \ref{tab:ident_param}. It can be seen the identified Lagrangian basis functions are exactly the same as the ground truth, and the system parameters are very close to the actual system.

\subsection{Example 6: Incompressible Navier-Stokes equation}\label{example_ns}
In this example, we extend the performance study to two-dimensional continuous systems. In particular, we consider the nonlinear parabolic incompressible Navier-Stokes equation that plays a key role in modeling various scientific and engineering processes involving weather, ocean currents, blood flow, and airflow around a wing. In this example, we artificially simulate the turbulent channel flow using the incompressible Navier-Stokes equation under a constant force field and aim to discover the associated Lagrangian of the incompressible flow. To that end, we use the following equations to generate the synthetic data,
\begin{equation}\label{eq:ns}
    \begin{aligned}
        & \frac{\mathcal{D} \bm{u}}{\mathcal{D} t} = - \frac{1}{\rho} \nabla p + \nu \Delta \bm{u} + f \cdot \bm{I}, \; (x,y) \in [0,1], \; t \in [0,T], \\
        & \nabla \cdot \bm{u} = 0, \; (x,y) \in [0,1], \\
        & (\bm{u}, \nabla p)(x, \partial y, t) = 0, \\
        & (\bm{u}, p)(x=0,y,t) = (\bm{u}, p)(x=1,y,t), \; \\
        & (\bm{u}, p)(x,y,t=0) = 0.01, \; (x,y) \in [0,1],
    \end{aligned}
\end{equation}
where ${\mathcal{D} \bm{u}}/{\mathcal{D} t} = \partial_t \bm{u} + (\bm{u} \cdot \nabla)\bm{u}$ is the material derivative, $\bm{u} \in \mathbb{R}^2$ is the vector-valued fluid velocity field, $p \in \mathbb{R}$ is the scalar pressure field, $\rho \in \mathbb{R}_{>0}$ is the fluid density, $\nu = \mu / \rho$ is the kinematic viscosity, and $\mu \in \mathbb{R}_{>0}$ is the dynamic viscosity. The force field is modeled as $f(x,y) = 0.1(\operatorname{sin}(2\pi(x + y)) + \operatorname{cos}(2\pi(x + y)))$. $\bm{I}=[1,0]^{\top}$ is the influence vector indicating the location of source $f(x,y)$, and $(\cdot)$ is the Schur product. For synthetic data simulation, $\nu=10^{-4}$ and  $\rho=800$kg/m$^3$ is considered. Simulation is performed for $T=1$s using $\Delta t$=0.001s on a $\mathbb{R}^{128 \times 128}$ spatial field. Contrary to the previous continuous examples, instead of discovering the Lagrangian density at each discretized location, we reshape the observation matrix into a vector and discover the Lagrangian of the complete system. 
\textbf{Results}: The basis functions that are accepted with a PIP$>$0.5 are displayed in Fig. \ref{fig:stem_basis}. Similar to the previous example, we observe that the proposed algorithm obtains a sparse and parsimonious representation of the parameter vector $\bm{\theta} \in \mathbb{R}^{15}$, which originally contains 15 regression coefficients. The identified kinetic energy functions is $\mathcal{T} = \bm{u}^2$, and the potential energy function is $\mathcal{V}=-\rho^{-1}p+\nu\nabla \bm{u}+f\cdot\bm{I}x$. The final form of the Lagrangian density is given in Table \ref{tab:ident_param}. The relative error clearly evidences the accuracy of the identified model.

\subsection{Learning analytical expression of Hamiltonian}
\begin{figure}[ht!]
    \centering
    \includegraphics[width=0.6\textwidth]{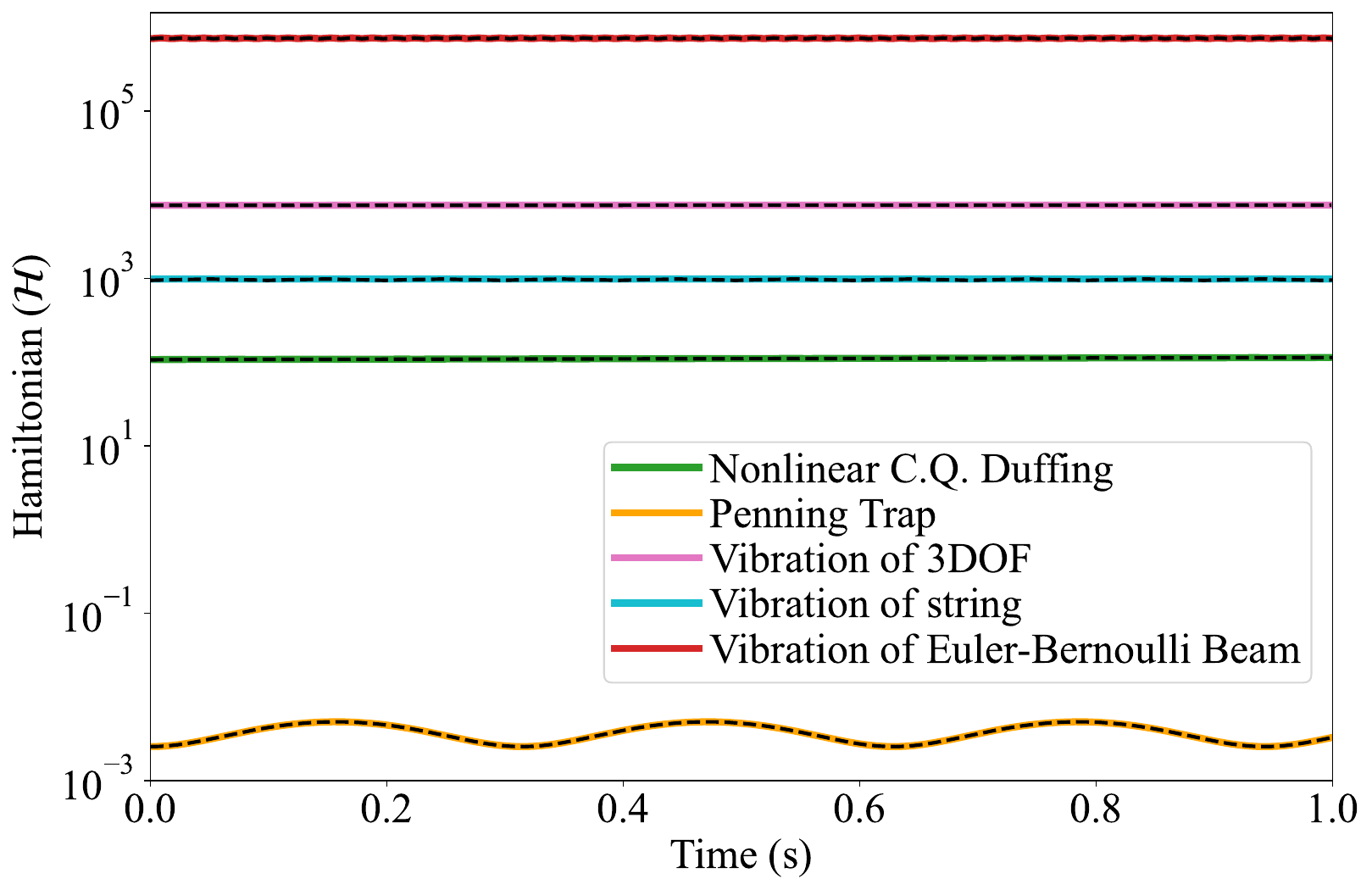}
    \caption{{Hamiltonian discovery of example problems}. The solid lines represent the actual energy of the underlying systems, and the dotted lines denote the Hamiltonian of the identified systems. In all the examples, the conservation of energy is discovered. The total energy discovered using the proposed framework matches the Hamiltonian of the actual system.}
    \label{fig:energy}
\end{figure}

The proposed sparse Bayesian framework discovers the exact analytical form of the underlying Lagrangian. Therefore, the direct application of the Legendre transformation on the discovered Lagrangian model yields the interpretable form of the associated Hamiltonian. The Hamiltonian of an underlying system from the discovered Lagrangian is derived using the Eq. \eqref{eq:hamiltonian}. The learned Hamiltonians of the systems are provided in Table \ref{table:identify}. For comparison, the discovered Hamiltonians of the systems are compared with the total energy of the system (i.e., $\mathcal{H} = \mathcal{T} + \mathcal{V}$). The results are shown in Fig. \ref{fig:energy}, where we observe that the discovered Hamiltonians perfectly overlap the ground truth and remain stationary for the entire duration of time. The proposed framework also learns the uncertainties associated with the discovered Hamiltonian model; however, the confidence intervals are not visible in the given figure due to the small magnitudes of the uncertainties. Overall, the results suggest that the proposed framework also eliminates the need for a separate framework for the Hamiltonian discovery of dynamical systems.
\begin{table}[!t]
\centering
\begin{threeparttable}
\centering
\caption{Learning Energy conservation (Hamiltonian) and ODE/PDE descriptions}
\label{table:identify}
\begin{tabular}{lll}
    \toprule
    \textbf{Systems} & \multicolumn{2}{l}{\textbf{Hamiltonian and ODE/PDE descriptions}} \\
    \midrule
    Cubic–quintic Duffing & $\mathcal{H}^{*}$: & $\frac{1}{2}\dot{X}^2 + \underset{\textcolor{blue}{\pm 1.50}}{498.46}X^2 + \underset{\textcolor{blue}{\pm 21.48}}{1226.76}X^4 + \underset{\textcolor{blue}{\pm 92.11}}{14912.12}X^6$ \\
    & ODE: & $\ddot{X}(t) + \underset{\textcolor{blue}{\pm 3.00}}{996.92} {X}(t) + \underset{\textcolor{blue}{\pm 42.96}}{2453.52} {X}^{3}(t) + \underset{\textcolor{blue}{\pm 187.22}}{29824.24} {X}^{5}(t) = 0$ \\
    \midrule
    Penning Trap & $\mathcal{H}^{*}$: & $\frac{1}{2}(\dot{X}^2 + \dot{Y}^2+ \dot{Z}^2) + 24.99{\scriptstyle\textcolor{blue}{\pm 0.01}}(X^2 + Y^2 - 2Z^2)$ \\
    & ODE: & $\ddot{X}(t) = 99.97{\scriptstyle\textcolor{blue}{\pm 1.32}}\dot{Y}(t) + 49.99{\scriptstyle\textcolor{blue}{\pm 0.02}}X(t)$, \\
    && $\ddot{Y}(t) = -99.97{\scriptstyle\textcolor{blue}{\pm 1.32}}\dot{X}(t) + 49.99{\scriptstyle\textcolor{blue}{\pm 0.02}} Y(t)$, \\
    && $\ddot{Z}(t) = - 49.99{\scriptstyle\textcolor{blue}{\pm 0.02}} Z(t)$ \\ 
    \midrule
    Vibration of Structure & $ \mathcal{H}^{*}$: & $\frac{1}{2} \sum_{k=1}^{3} \dot{X}_k^2 + ( \underset{\textcolor{blue}{\pm 0.18}}{2497.68} x_1^2 + \underset{\textcolor{blue}{\pm 0.23}}{2497.67}(X_2-X_1)^2 + \underset{\textcolor{blue}{\pm 0.41}}{2497.14} (X_3-X_2)^2 )$ \\
    & ODEs: & $\ddot{X}_1 + {4995.36}{\scriptstyle \textcolor{blue}{\pm 0.36}}{X_1} + {4994.39}{\scriptstyle \textcolor{blue}{\pm 0.46}}(X_1-X_2) = 0$ \\
    && $\ddot{X}_2 + {4997.33}{\scriptstyle \textcolor{blue}{\pm 0.30}}(X_2-X_1) + {4995.88}{\scriptstyle \textcolor{blue}{\pm 0.28}}(X_2-X_3) = 0$\\
    && $\ddot{X}_3 + {4994.28}{\scriptstyle \textcolor{blue}{\pm 0.82}}(X_3-X_2) = 0$\\
    \midrule
    Vibration of string & $\mathcal{H}^{*}$: & $\frac{1}{2}\sum_{i} \dot{u}_i^2 + \sum_i {9.98}{\scriptstyle \textcolor{blue}{\pm 0.05}} (\partial_x u_{i})^2$ \\
    & PDE: & $\partial_{tt} u(X,t) - ({9.98}{\scriptstyle \textcolor{blue}{\pm 0.05}})^2 \partial_{xx} u(X,t) = 0$ \\
    \midrule
    Vibration of beam &  $\mathcal{H}^{*}$: & $\frac{1}{2} \sum_{i} \dot{u}_i^2 + {2.1168}{\scriptstyle \textcolor{blue}{\pm 0.01}} \sum_i (\partial_{xx} u_{i})^2$ \\
    & PDE: & $\partial_{tt} u(X,t) - ({2.1168}{\scriptstyle \textcolor{blue}{\pm 0.01}}) \partial_{xxxx} u(X,t) = 0$ \\
    \bottomrule
\end{tabular}
\end{threeparttable}
\end{table}

\subsection{Governing ODE/PDE descriptions from Lagrangian}
\begin{figure}[ht]
    \centering
    \includegraphics[width=\textwidth]{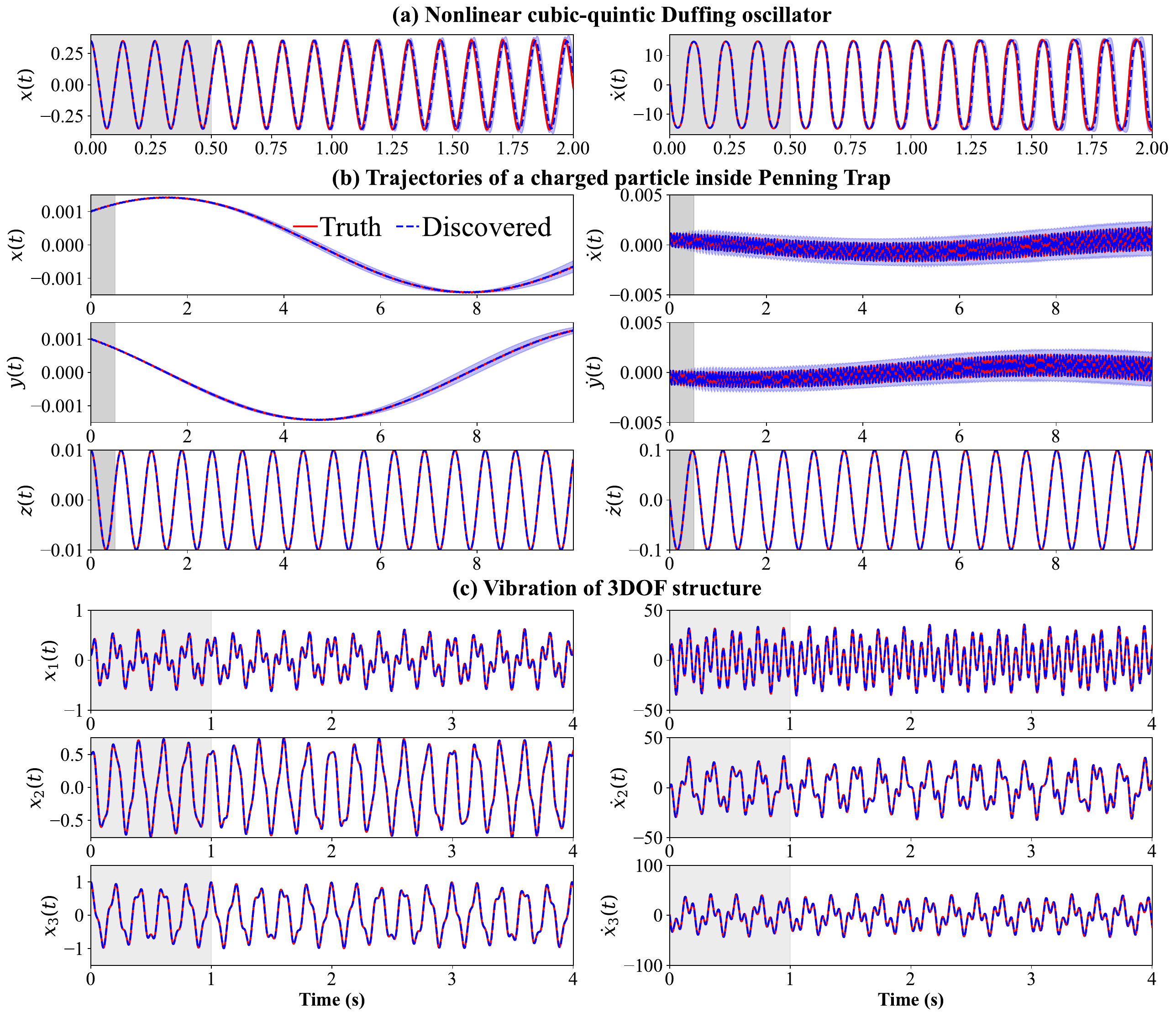}
    \caption{Predictive performance of the identified Lagrangian equations of motion of discrete systems. The grey shaded region indicates the training period, and the blue response bound indicates the confidence interval of the solution.}
    \label{fig:response_lin_nonlin}
\end{figure}

In this section, we utilize the Euler-Lagrange equation on the identified Lagrangian for distilling equations of motion of underlying systems. Since the proposed framework provides an exact analytical description of the Lagrangian, the Euler-Lagrangian equation directly provides the interpretable forms of equations of motion. However, as compared to the previous data-driven equation discovery techniques \cite{brunton2016discovering,nayek2021spike,tripura2023sparse}, where the governing equations of motion are discovered using the first-order equations, the proposed framework directly yields the second-order motion equations. The resulting ODE/PDE descriptions and the model uncertainties are provided in Table \ref{table:identify}. The predicted responses using the discovered models for the Nonlinear cubic-quintic oscillator, the Penning trap, and the 3DOF structure are shown in Fig. \ref{fig:response_lin_nonlin}. The training data regime is illustrated by the grey shaded area. The response bounds indicate the 95\% confidence interval averaged over 100 random seeds. In all three cases, we observe that the mean of the predicted responses almost exactly approximates the ground truth in both the training data regime and the prediction regime. The predictive uncertainty in the predictions is illustrated by the narrow width of the 95\% confidence further dictates the accuracy of the identified model. 

\begin{figure}[!ht]
    \centering
    \includegraphics[width=\textwidth]{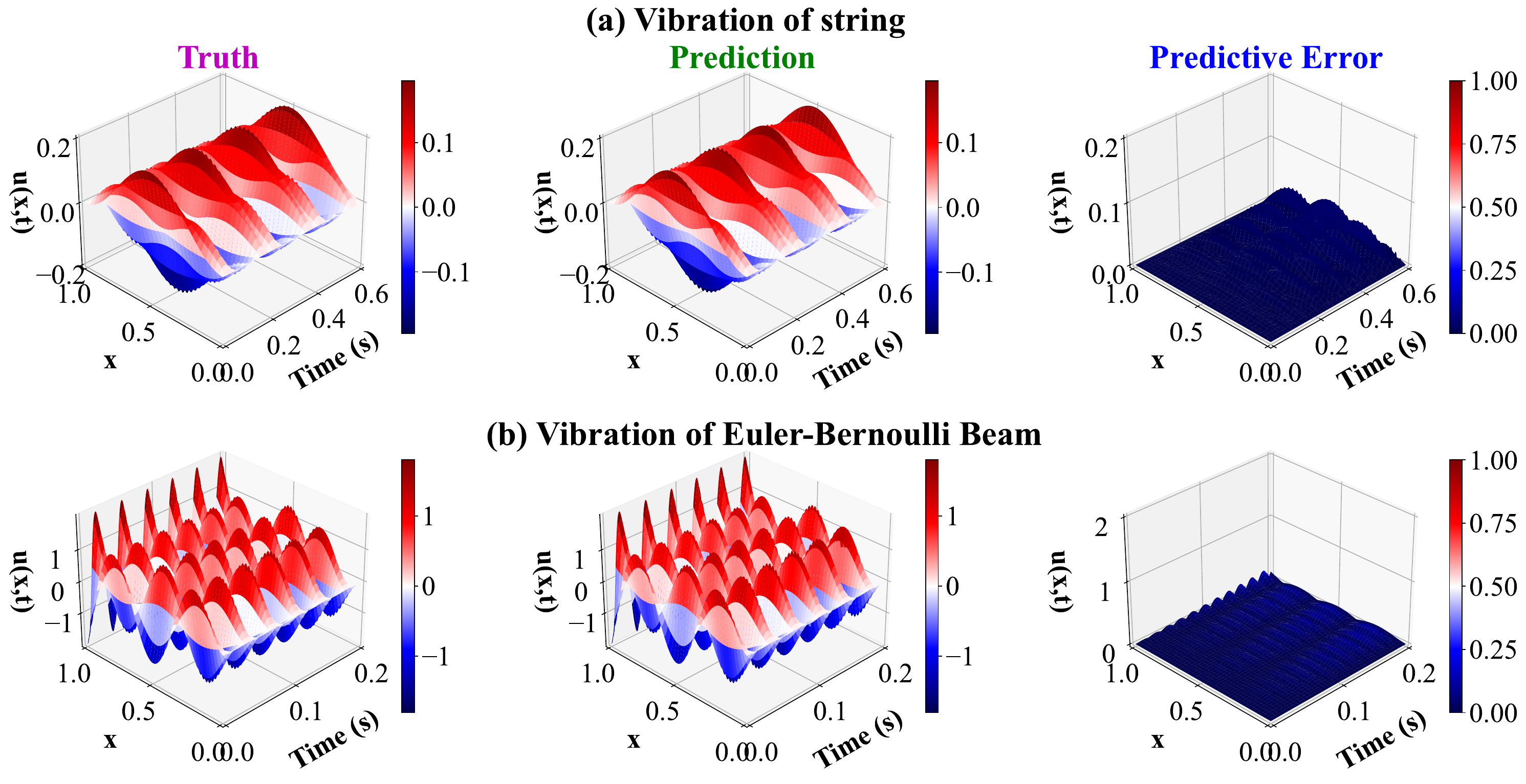}
    \includegraphics[width=\textwidth]{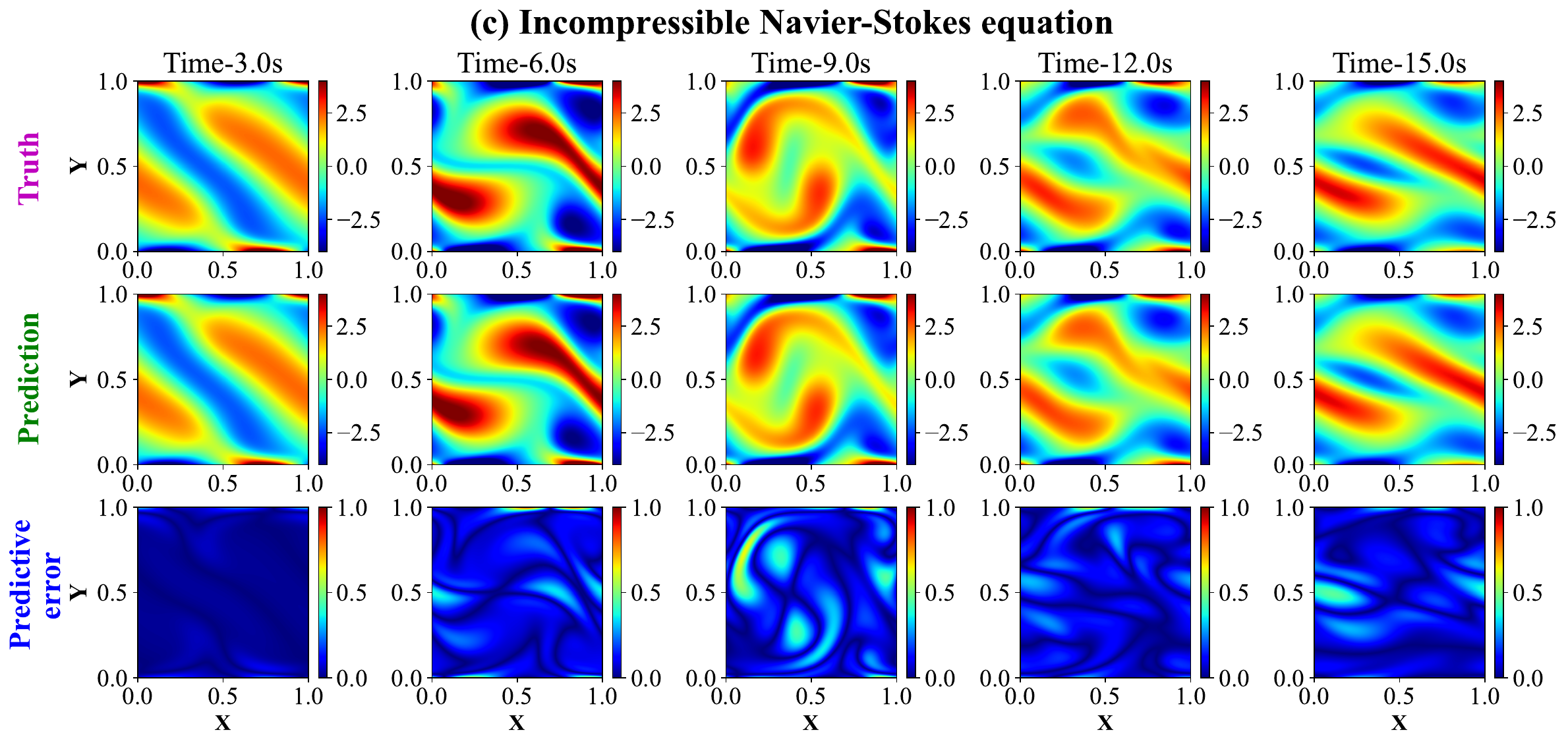}
    \caption{{Predictive performance of the Lagrangian equations of motion of the String, Euler-Bernoulli Beam, and incompressible Navier-Stokes equation.}. The solutions of the actual systems and identified Lagrangian equations of motion are compared.}
    \label{fig:Response_pde}
\end{figure}

The ground truth, the responses predicted using the identified model, and the associated error between them for the continuous system, i.e., the string, the Euler-Bernoulli beam, and the incompressible Navier-Stokes equation, are shown in Fig. \ref{fig:Response_pde}. The graphical representations of the system responses and the predictive error plots evidence the accuracy of the identified models. The ability to discover the fourth-order PDE governing the transversal motion of the Euler-Bernoulli beam and complex nonlinear functions like divergence and Laplacian in incompressible Navier-Stokes suggest that the proposed algorithm can accurately discover complex and high-order mathematical models entirely from data. Further, it can be noticed that even with the 1s data (with 1000Hz sampling frequency), the discovered Navier-Stokes model can predict accurate solutions up to 15s. Overall, the results suggest that the proposed framework can distill the Lagrangian of both discrete (i.e., ODEs) and continuous systems (i.e., PDEs). The learned Lagrangian meets the exact analytical descriptions of the underlying systems and is accurate enough to provide exact analytical equations of motion from data.

\subsection{Generalization to high-dimensional systems and long-term prediction}
When data are collected from many sensor locations (e.g., spatial discretization of PDEs, number of DOF in ODEs), the state dimension $m$ becomes prohibitively large. In these cases, discovering the Lagrangian may become computationally expensive due to the exponential growth of the dictionary dimension. However, using the proposed framework, we can identify the underlying Lagrangian from a subset of the measurements and then analytically extend to $m$-dimensions. For demonstration, we consider the Lagrangian of the structure and generalize it to a 100-DOF system.
The Lagrangian of the multi-dimensional structure follows,
\begin{equation}
    \mathcal{L} = \sum_{j=1}^{n=100} \frac{1}{2}\dot{X}_j^2 - \underset{\textcolor{blue}{\pm 0.18}}{2497.68} X_1^2 - \sum_{j=1}^{n=99} \left({   \underset{\textcolor{blue}{\pm 0.32}}{2497.41}(X_{j+1}-X_{j})^2 }\right),
\end{equation}
where the parameter values are obtained by averaging the identified parameters of the 3DOF structure.
Further, in all scientific machine learning applications, the prime focus is to distill the actual dynamics of the underlying system rather than to predict the observed data. 
\begin{figure}[!ht]
    \centering
    \includegraphics[width=\textwidth]{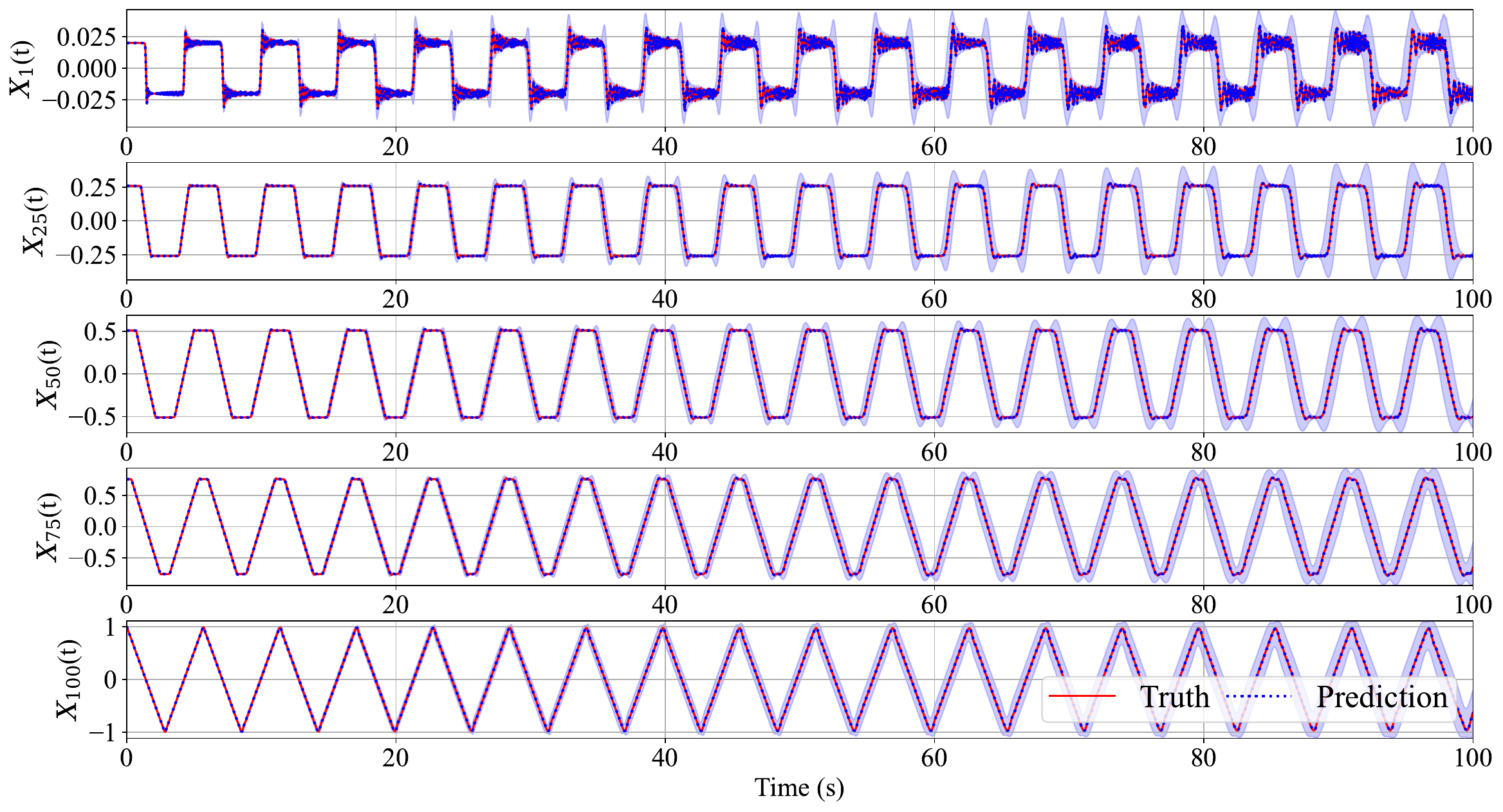}
    \caption{High-dimensional generalization ability of the proposed framework. The discovered 3DOF structural system is generalized to a 100-dimensional structural model. The shaded region indicates the 95\% confidence interval averaged over 100 random seeds.}
    \label{fig:case_general}
\end{figure}
To test the predictive ability of the generalized system, we have performed a long-term prediction of the system responses for $T$=100s. We use a different initial condition, where a linear displacement profile with $X_1$=0.1 and $X_{100}$=1. The results are summarized in Fig. \ref{fig:case_general}. The responses of the generalized system evolve indistinguishably along with the solutions of the actual system. Therefore, it is evident that even with the measurement of 1s, the proposed framework could learn the exact description of the Lagrangian and generalize it to a high-dimensional system whose predictive ability does not wear off rapidly.

\begin{table}[!ht]
    \centering
    \caption{Noise sensitivity of the proposed framework}
    \begin{tabular}{llllll}
        \toprule
         Noise level $\zeta$ (\%) & \multicolumn{4}{c}{Relative $L^2$ error (\%)} \\ \cline{2-5}
         & C.Q. Duffing & Penning Trap & 3DOF & Vibration of String \\
         \midrule
         No noise & 0.6037 & 0.0912 & 0.1006 & 0.1900 \\
         \hline
         2 & 2.7251 & 0.3910 & 20.4329 & 1.7833 \\
         5 & 7.9100 & 1.0934 & -- & 8.6631 \\
         10 & 15.5811 & 3.3225 & -- &  25.0291 \\
         15 & 20.4136 & 16.5744 & -- & -- \\
         \bottomrule
    \end{tabular}
    \label{tab:noise_error}
\end{table}
\subsection{Sensitivity to measurement noise}
In this section, we test the performance of the proposed framework against the measurement noise, where we corrupt the single observation of data with zero mean Gaussian noise. A total of four levels of noise are considered, where each level indicates a certain percentage of the standard deviation of the actual data. In the comparison, only those cases where the correct dictionary basses are identified are reported. If the identified sparse model exactly matches the description of the Lagrangian of the actual system, we compute an error $e_{\mathcal{L}} = 100 \times \|\mathcal{L}_a - \mathcal{L}^* \|_2 / \|\mathcal{L}_a\|_2$, with $\mathcal{L}_a$ and $\mathcal{L}^*$ representing the actual and identified Lagrangian, respectively. The corresponding results are illustrated in Table \ref{tab:noise_error}. 
We observe that in the presence of noise, the proposed framework fails to correctly identify the analytical description of the Lagrangian for the 3DOF system. For the nonlinear cubic-quintic Duffing oscillator, the proposed framework yields a relative $L^2$ error of $\leq 5\%$ for noise levels $\zeta \leq 5\%$.
Similarly, for the Penning trap example and the string vibration, the proposed framework yields an error $\leq 5\%$ when the noise level $\zeta$ is $\leq 15\%$ and $\leq 5\%$, respectively.  The study concludes that the proposed framework works successfully when the data are corrupted at a very mild level of noise, i.e., $\zeta \leq 2-5\%$.

\begin{figure}[!t]
    \centering
    \begin{subfigure}[b]{0.32\textwidth}
        \centering
        \includegraphics[width=\textwidth]{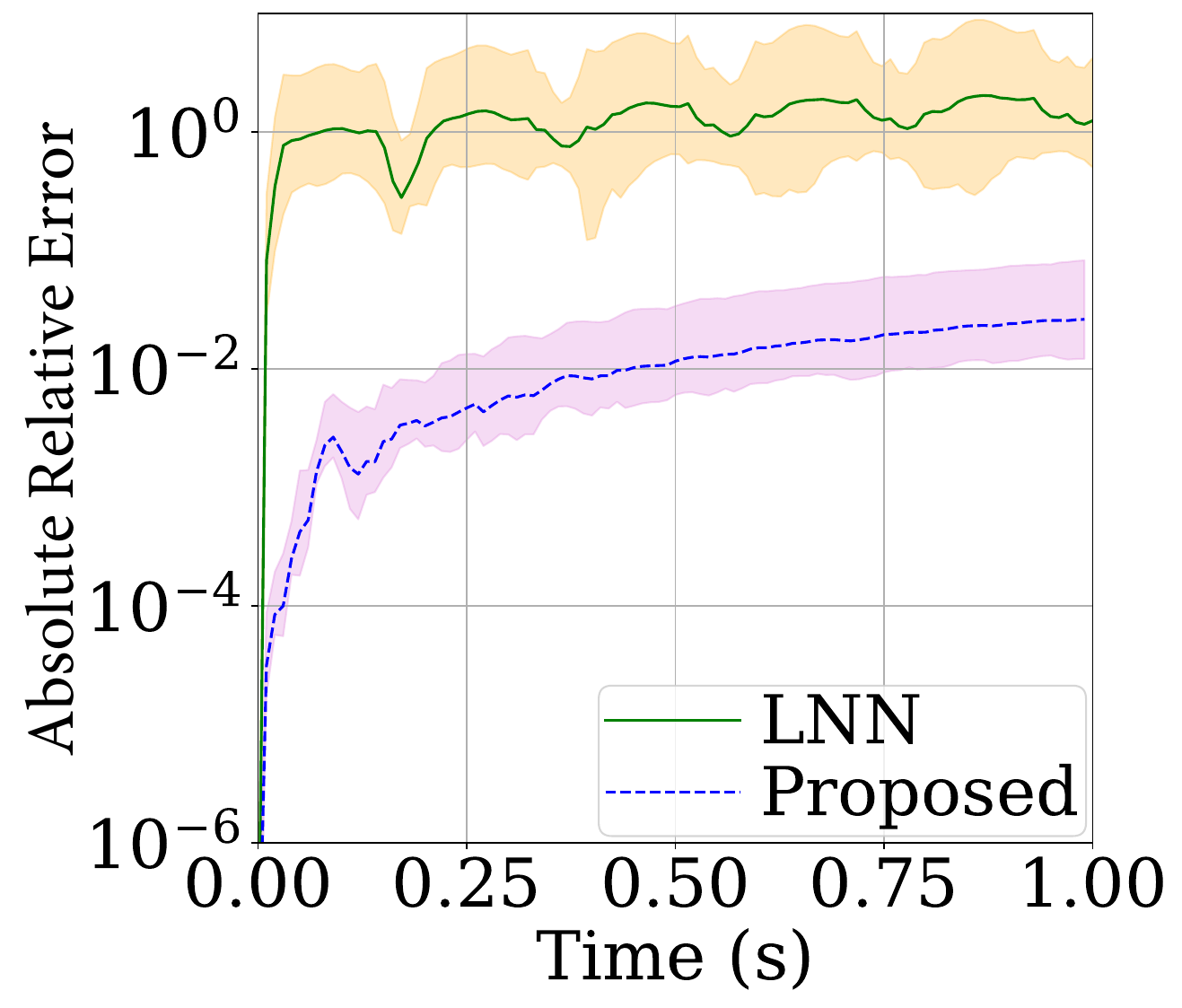}
        \caption{Cubic-quintic Duffing}
        \label{fig:Lnn_error_harmon}
    \end{subfigure}
    \begin{subfigure}[b]{0.32\textwidth}
        \centering
        \includegraphics[width=\textwidth]{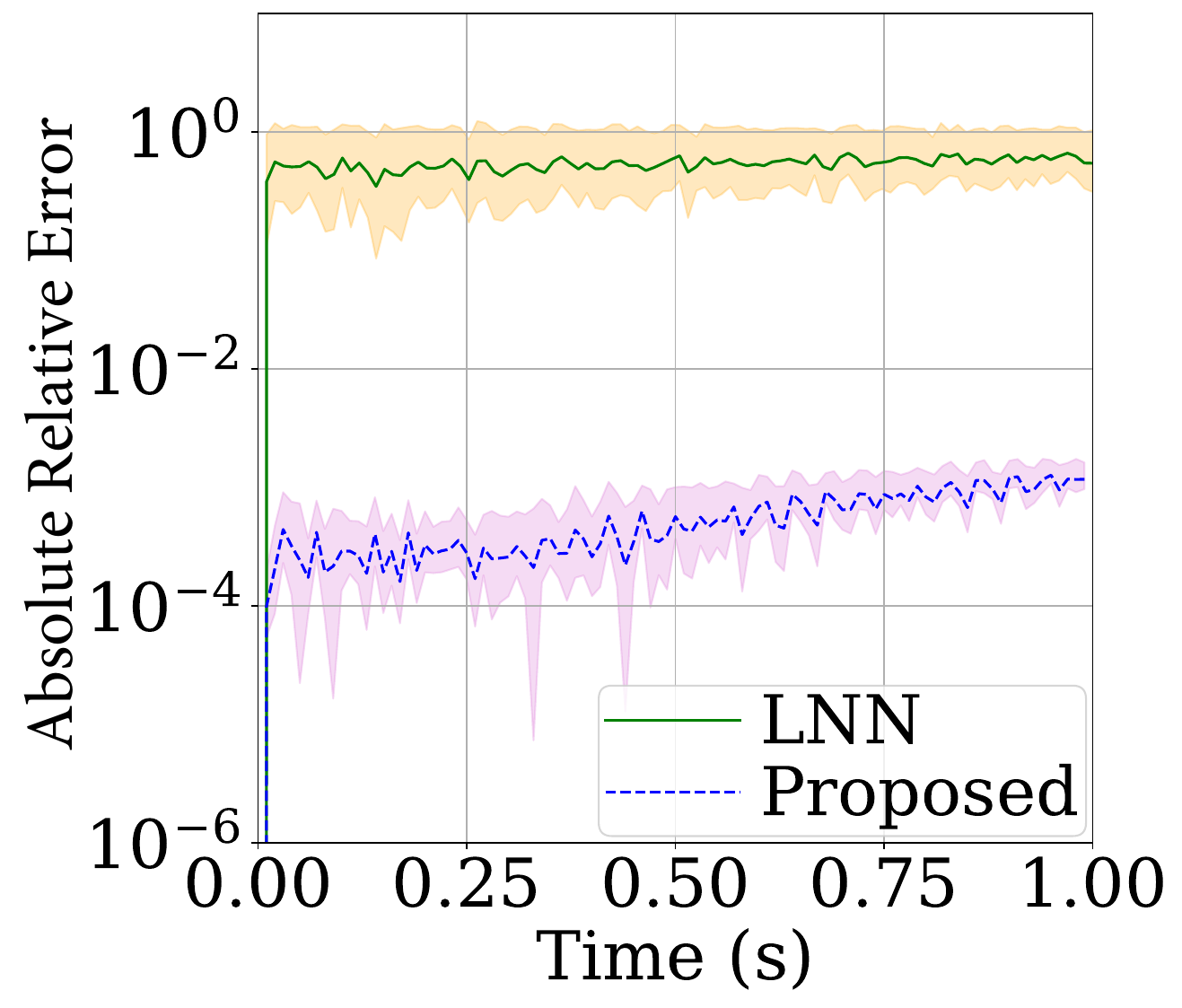}
        \caption{Penning Trap}
        \label{fig:Lnn_pred_harmon}
    \end{subfigure}
    \begin{subfigure}[b]{0.32\textwidth}
        \centering
        \includegraphics[width=\textwidth]{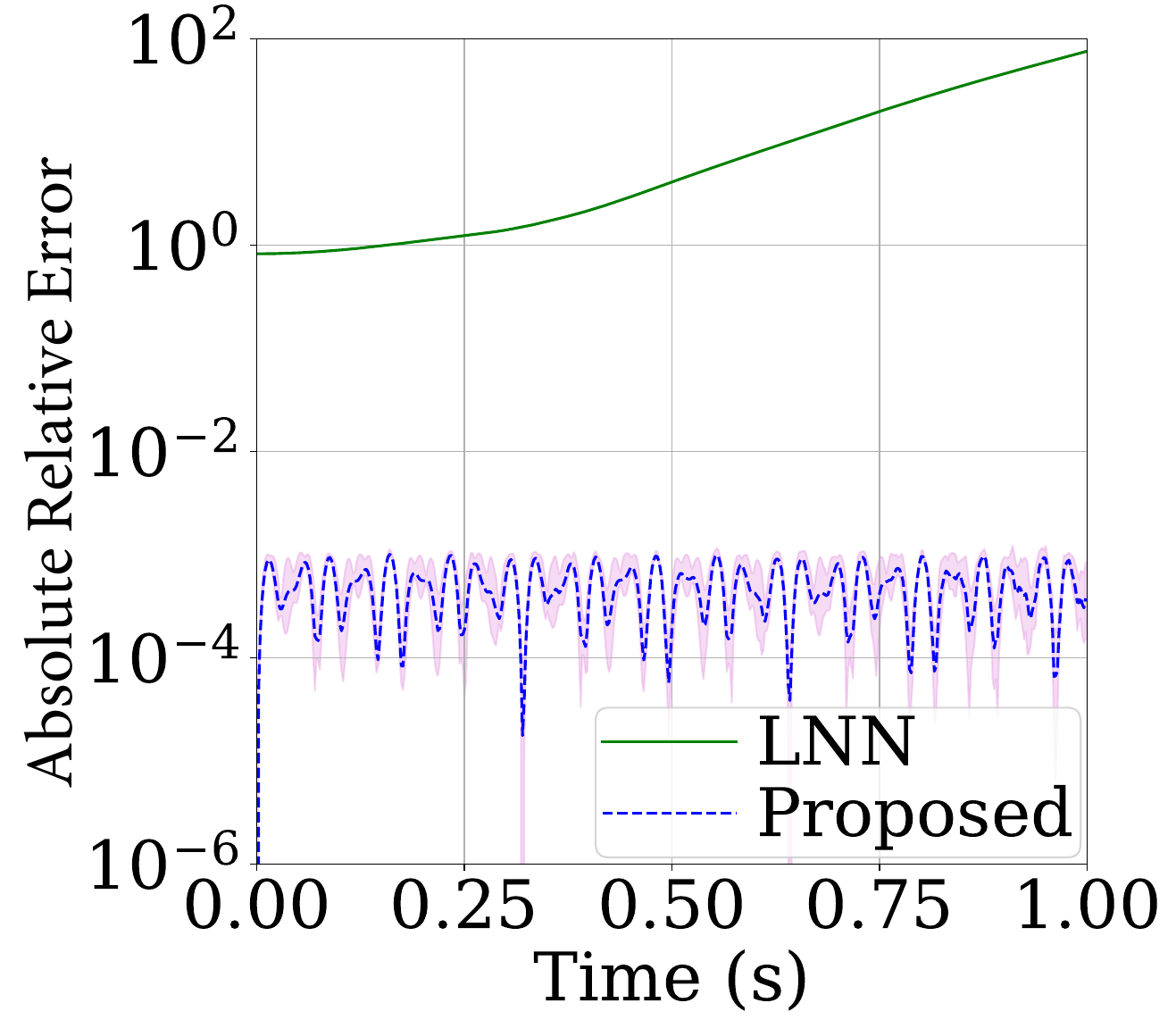}
        \caption{3DOF structure}
        \label{fig:Lnn_error_3dof}
    \end{subfigure}
    \caption{Predictive performance of the proposed Bayesian algorithm against LNN. The shaded region denotes the 95\% confidence interval from 25 different initial conditions. The mean values of the identified (using the proposed Bayesian algorithm) system parameters are considered for performing predictions.}
    \label{fig:Lnn_harmon}
\end{figure}
\subsection{Comparative study:}\label{sec:comparative}
This section compares the Hamiltonian derived from the proposed Bayesian algorithm with that obtained from the recently proposed Lagrangian neural network (LNN) \cite{cranmer2020lagrangian}. The comparative results for the nonlinear cubic-quintic Duffing oscillator, the Penning trap, and the 3DOF structure are provided in Fig. \ref{fig:Lnn_harmon}. In these three examples, we observe that the proposed algorithm almost exactly conserved the true energy of the system. In contrast, the energy violation in the LNN model is comparatively very high. We have also estimated the relative $L^{2}$ norm of the Hamiltonian discrepancies for both the proposed and LNN discovered models with respect to the actual total energy. Averaged over 25 different initial conditions, the relative $L^2$ norm of the discrepancies for the cubic-quintic Duffing oscillator, Penning trap, and 3DOF structural system were found to be 0.06\%, 0.33\%, and 0.06\% for the proposed algorithm and 60.25\%, 57.23\%, and 89.95\% for the LNN models.

\section{Discussions}\label{sec:conclusion}
A sparse Bayesian identification framework for discovering parsimonious and interpretable models of Lagrangian from data is introduced. The overarching goal of discovering interpretable Lagrangian densities from data is formulated as a problem of identifying Lagrangian functions from a pre-defined physics-based library of candidate basis functions. Instead of directly discovering the Lagrangian from the Lagrangian library, the sparse regression is performed on the Euler-Lagrange library, which is discovered by operating the Euler-Lagrangian operator on the Lagrangian library, thereby motivating the discovered Lagrangian model to satisfy the optimal path condition. At the core of sparse regression, the proposed framework utilizes the discontinuous spike-and-slab prior for basis function selection cum parameter estimation problem. 
The novelty of the proposed framework lies in the fact that it automates the distillation of interpretable expressions for Lagrangian density, associated Hamiltonian, and the interpretable ODE/PDE descriptions of underlying physical systems in a single framework and in one-go. From learning Lagrangian to the distillation of Hamiltonian and ODE/PDE descriptions, the proposed framework utilizes only a single time-series measurement of system states. Being Bayesian, it also quantifies model-form uncertainties. The ability to learn interpretable Lagrangian, Hamiltonian, and governing equations of motion using a single observation is highly missing in the literature.

We explored the performance of the proposed Bayesian framework in a series of numerical experiments. The results indicated that the discovered Lagrangian models very closely emulate the ground truth for both the discrete and continuous systems. The models show robustness in correctly discovering the energy conservation in the underlying physical systems. The discovered equations of motion also show generalization to high-dimensional systems and long predictive ability, predicting the actual system dynamics for a duration greater than 100 times the training period. 
Lagrangian formalism is the starting point of all modern physics, from continuum mechanics to electromagnetism to quantum theory. Due to sensor advancements, we now have an abundance of data; however, the processing abilities of humans are limited. In such cases, the proposed framework can assist humans in accelerating the pace of discovering new physical laws. Through analytical descriptions of Lagrangian, it can provide reusable physical knowledge of physical systems, enabling the knowledge-based design of future technologies such as robotic control, turbulence mixing, and modeling of thermal transport in alloys.

One limitation of the proposed framework resides in the right selection of candidate functional forms in the design matrix. A possible solution is to include all the possible kinetic and potential energy functions in the candidate library and then use the basis functions' uncertainties to create a parsimonious model. However, it is yet unclear how to design a strategy to best select the most probable basis functions in the design matrix. A future endeavor also includes extending the proposed framework for learning the interpretable Lagrangian densities of dissipative and constrained systems.

\section*{Acknowledgements} 
T. Tripura acknowledges the financial support received from the Ministry of Education (MoE), India, in the form of the Prime Minister's Research Fellowship (PMRF). S. Chakraborty acknowledges the financial support received from Science and Engineering Research Board (SERB) via grant no. SRG/2021/000467, and from the Ministry of Port and Shipping via letter no. ST-14011/74/MT (356529).

\section*{Declarations}

% \subsection*{Funding} The corresponding author received funding from IIT Delhi in form of seed grant.

\subsection*{Conflicts of interest} The authors declare that they have no conflict of interest.

% \subsection*{Availability of data and material} Upon acceptance, all the source codes to reproduce the results in this study will be made available to the public on GitHub by the corresponding author.

\subsection*{Code availability} Upon acceptance, all the source codes to reproduce the results in this study will be made available to the public on GitHub by the corresponding author.

% Bibliography
% ~~~~~~~~~~~~~~~~~~~~~~~~~~~~~~~~~~~~~~~~~~

\end{document}